\documentclass[preprint, sort, compress, 12pt]{elsarticle}

\usepackage{graphicx} 
\graphicspath{{./plots}}
\usepackage{amsmath}
\usepackage{float}
\usepackage{lipsum}
\makeatletter
\def\ps@pprintTitle{%
 \let\@oddhead\@empty
 \let\@evenhead\@empty
 \def\@oddfoot{}%
 \let\@evenfoot\@oddfoot}
\makeatother

\usepackage[utf8]{inputenc}
\usepackage{algpseudocode}
\usepackage{algorithm}
\usepackage[margin=1in]{geometry}
\usepackage{bold-extra}
\usepackage[normalem]{ulem}
\usepackage{adjustbox}
\usepackage{appendix}
\usepackage[pagewise]{lineno}
\usepackage{diagbox}

\usepackage{amssymb}
\usepackage{amsthm}
\usepackage{times}
\usepackage[titles]{tocloft}
\usepackage{enumerate}
\usepackage{bm}
\usepackage{sidecap}
\usepackage{pdfpages}
\usepackage{soul}
\usepackage{multicol}
\usepackage{float}
\usepackage{multibib}
\usepackage{geometry}
\usepackage[small,compact]{titlesec}
\usepackage[normal,bf,labelsep=colon,skip=2pt]{caption}
\usepackage{setspace}
\usepackage{subfig}
\usepackage{tocloft}
\usepackage{url}
\usepackage{xurl}
\usepackage{floatflt}
\usepackage{array}
\usepackage{pgfgantt}
\usepackage{sidecap}
\usepackage{hyperref}
\usepackage{multirow}
\numberwithin{equation}{section}
\numberwithin{figure}{section}
\numberwithin{table}{section}

\newcommand*\diff{\mathop{}\!\mathrm{d}}

\begin{document}

\begin{frontmatter}
\title{Neural Dynamical Operator: Continuous Spatial-Temporal Model with Gradient-Based and Derivative-Free Optimization Methods}

\author{Chuanqi Chen}
\author{Jin-Long Wu\corref{cor1}} \ead{jinlong.wu@wisc.edu} 
\cortext[cor1]{Corresponding author}

\address{Department of Mechanical Engineering, University of Wisconsin–Madison, Madison, WI 53706}

\begin{abstract}
Data-driven modeling techniques have been explored in the spatial-temporal modeling of complex dynamical systems for many engineering applications. However, a systematic approach is still lacking to leverage the information from different types of data, e.g., with different spatial and temporal resolutions, and the combined use of short-term trajectories and long-term statistics. In this work, we build on the recent progress of neural operator and present a data-driven modeling framework called neural dynamical operator that is continuous in both space and time. A key feature of the neural dynamical operator is the resolution-invariance with respect to both spatial and temporal discretizations, without demanding abundant training data in different temporal resolutions. To improve the long-term performance of the calibrated model, we further propose a hybrid optimization scheme that leverages both gradient-based and derivative-free optimization methods and efficiently trains on both short-term time series and long-term statistics. We investigate the performance of the neural dynamical operator with three numerical examples, including the viscous Burgers' equation, the Navier--Stokes equations, and the Kuramoto--Sivashinsky equation. The results confirm the resolution-invariance of the proposed modeling framework and also demonstrate stable long-term simulations with only short-term time series data. In addition, we show that the proposed model can better predict long-term statistics via the hybrid optimization scheme with a combined use of short-term and long-term data.
\end{abstract}

\end{frontmatter}

\section{Introduction}

Complex dynamical systems are ubiquitous in many science and engineering applications, e.g., mixing phenomena in atmospheric and ocean dynamics~\cite{gill1982atmosphere,fernando1991turbulent,vallis2017atmospheric}, physical and chemical processes of energy conversion~\cite{dimotakis2005turbulent}, and drones operating in turbulent flows~\cite{shi2019neural}. For most of these systems, numerically simulating the governing equations derived by first principles is still infeasible in the foreseeable future, and thus closure models are inevitably needed to account for those unresolved degrees of freedom. Many of the classical closure models are calibrated with simplified settings and are known to have limited predictive capability, mainly due to the lack of enough expressive power in the model form and the empirical calibration to a small amount of data. In the past few decades, we have witnessed the rapid growth of high-fidelity simulation and experimental data and the great advance of machine learning methods, which motivated the development of data-driven modeling techniques~\cite{brunton2016discovering,kutz2016dynamic,wang2017physics,wu2018physics,duraisamy2019turbulence,brunton2020machine,schneider2021learning,gupta2021neural,brunton2022data,chen2023causality,chen2023ceboosting, chen2024cgnsde, dong2024data} with the aim of improving or even replacing the classical models by neural-network-based models.

Standard machine learning methods have achieved great success in the past decade, e.g., convolutional neural networks~\cite{lecun1989handwritten,lecun1998gradient} for tasks such as computer vision (CV), and recurrent neural networks~\cite{funahashi1993approximation,hochreiter1997long,jaeger2001echo,maass2002real} for tasks such as natural language processing (NLP). More recently, transformer-based models have demonstrated even better performance in both CV and NLP tasks. Inspired by the success of these methods in standard machine learning tasks, many previous works explored using these methods to improve the modeling of dynamical systems. Despite the initial promising results, standard machine learning methods often assume fixed discretization in both space and time, which presents a significant constraint for accurately modeling complex dynamical systems. The main reason for such a limitation is two-fold: (i) the available data sources are unlikely in a consistent resolution, and (ii) the efficient characterization of the system state usually demands a non-uniform resolution. For instance, the data sources of earth system include high-fidelity simulations~\cite{wedi2020baseline} with a fine spatial resolution such as $O(100\textrm{m})$ to $O(1\textrm{km})$, satellite images across a wide range of resolutions with the finest one of $O(10\textrm{m})$, and the data collected by observatories that sparsely distributed across the earth. The temporal resolutions also vary among high-fidelity simulations, satellite images, and observatory data. This multiscale nature of different data sources is also present in many engineering applications, e.g., wind farms with both LiDAR data and high-fidelity simulations of atmospheric boundary layer~\cite{porte2011large,lu2011large}. On the other hand, the system states of complex dynamics often have multiscale features intrinsically, for which an adaptive resolution is often the more efficient way of characterizing the system state than a uniform resolution.

The multiscale nature of both the system state and the available data sources of complex dynamical systems has motivated the development of continuous form machine learning methods for the modeling and simulation of those systems. For instance, Fourier neural operator (FNO)~\cite{li2021fourier} employs Fourier and inverse Fourier transformations to construct Fourier layers, which are used as building blocks to approximate an operator (i.e., a mapping between Banach spaces) in the continuous form. Essentially, FNO belongs to a general category of neural operators~\cite{kovachki2023neural}, i.e., using neural networks to approximate the integral operator as the basic component of the approximation of a more general operator. Other neural operator methods that belong to this category are graph neural operators (GNO)~\cite{anandkumar2020neural,li2020multipole} and low-rank neural operators (LNO), which connects to another popular neural operator framework known as DeepONet~\cite{lu2019deeponet,lu2021learning}. More specifically, DeepONet introduced branch net and trunk net to implement an approximation of operator based on the universal approximation theorem of operators. There are also other recent works that construct neural-network-based operators via a wavelet-based approximation of integral operators~\cite{gupta2021multiwavelet} or reconstruction from other types of sub-networks~\cite{patel2022variationally}. Comparisons between neural operator approaches have been studied in~\cite{lu2022comprehensive,kovachki2023neural}. The approximation error has been studied in~\cite{cao2023residual} with a residual-based error correction method being proposed. Physics constraints~\cite{li2021physics,wang2021learning} and derivative information~\cite{o2022derivate} were demonstrated as additional information sources that can enhance the performance of neural operator methods. There have been many other interesting extensions of the standard neural operator frameworks, e.g., solving problems on general geometries~\cite{li2022fourier}, learning of nonlinear non-autonomous dynamical systems~\cite{lin2023learning}, and improving neural operator techniques with the recent success of large language models~\cite{yang2023context,liu2023prose}, to name a few. Although these methods can be used to characterize the solution operator of a dynamical system, they only work with a fixed and uniform temporal resolution if standard neural operator methods are used. By taking the prediction lead time as an additional input, the neural operator methods can potentially work with non-uniform temporal resolution but often demand abundant data with different temporal resolutions for the training.

On the other hand, neural ordinary differential equation (ODE)~\cite{chen2018neural} provided a temporally continuous framework of machine learning methods for the modeling and simulation of dynamical systems. Recently, the combined use of neural ODE and neural operator has been explored by~\cite{cho2022neural} in the context of standard machine learning tasks such as classification and computer vision. In terms of spatial-temporal modeling of dynamical systems, the key relevant concepts of neural ODE are: (i) neural networks can be used to characterize the unknown vector field of a finite-dimensional dynamical system, and (ii) the unknown coefficients of the neural network can be trained via gradient descent using either the adjoint method or automatic differentiation. The idea of modeling continuous dynamics via machine learning methods was also discussed in~\cite{weinan2017}. In the past few years, several works~\cite{maulik2020time,portwood2019turbulence,linot2022data,djeumou2022neural} have explored the use of neural ODE in modeling dynamical systems and demonstrated promising results. Although neural ODE provides the flexibility of handling data with a non-uniform resolution in time, it has been shown in~\cite{linot2023stabilized} that the use of a standard network in neural ODE can lead to long-term instability, mainly due to the amplification of high wavenumber components, when applying the trained network and simulating the modeled dynamical system. To address the long-term stability, the unknown vector field was modeled by linear and nonlinear parts with standard neural networks in~\cite{linot2023stabilized} and the training via neural ODE demonstrated more stable long-term simulations. In this work, we show that the stable long-term simulations of neural ODE models can alternatively be achieved by filtering out some high wavenumber components through the construction of a neural dynamical operator, for which the Fourier neural operator~\cite{li2021fourier} serves the purpose nicely.

Although neural ODE provides an efficient tool for training a model to match the short-term behavior of an unknown dynamical system, training the model to also quantitatively match the long-term system behavior still has some challenges. One challenge is on the computational side, that the long-term training would require huge memory costs~\cite{paszke2017automatic,baydin2018automatic,margossian2019review} via backpropagation or potential stability issues in the backward-in-time simulation via the adjoint method. Another challenge is on the problem formulation side, that the long-term system behavior tends to be more sensitive to the small changes in the model, especially for chaotic systems or systems with discontinuities, making the standard adjoint method sometimes even infeasible~\cite{wang2014least}. These challenges motivate us to explore an alternative optimization approach that is both efficient and robust for matching the long-term behaviors between the model and the true dynamical system. More specifically, we choose the ensemble Kalman inversion (EKI) method that was proposed in~\cite{iglesias2013ensemble}. Unlike the backpropagation or the adjoint method that is designed for the gradient descent optimization, ensemble Kalman inversion is derivative-free, parallelizable, and robust with noises of the data and chaos or uncertainties of the system~\cite{schneider2021learning,wu2023learning}. In the past decade, many developments have been made to enhance Kalman inversion methods, both in theory~\cite{schillings2017analysis,ding2021ensemble,calvello2022ensemble} and in algorithms, such as various types of regularizations (e.g., linear equalities and inequalities~\cite{albers2019ensemble}, Tikhonov~\cite{chada2020tikhonov}, sparsity~\cite{schneider2022ensemble}, and $\ell_p$~\cite{lee2021l_p}, and known physics~\cite{zhang2020regularized}), uncertainty quantification with Langevin dynamics for sampling~\cite{garbuno2020interacting}, and using other types of Kalman filters~\cite{huang2022iterated}. More recently, the ensemble Kalman inversion was also explored in~\cite{bottcher2023gradient} for the training of neural ODE. Instead of merely using EKI to train a neural ODE, we investigate a hybrid optimization approach that uses the standard gradient-based method for short-term trajectory matching and the EKI method for long-term statistics matching.

In summary, we develop a spatially and temporally continuous framework called neural dynamical operator for the data-driven modeling of dynamical systems with spatial fields as system states, by leveraging the success of neural operator and neural ODE. Compared with classical methods, the proposed framework is more capable of capturing the underlying dynamics from data for both short-term state prediction and long-term statistics matching with resolution-invariant feature. The framework is tested with three dynamical systems, including 1-D Burgers' equation, 2-D Navier--Stokes equation, and Kuramoto--Sivashinsky equation. The first two examples focus on the performances of neural dynamical operator trained with short-term data, to demonstrate the merits of the neural dynamical operator in terms of (i) spatial-temporal resolution-invariant and (ii) stable long-term simulations. The third example demonstrates the merit of a hybrid optimization method that leverages both short-term and long-term data to calibrate the neural dynamical operator. The key contributions of this work are summarized below:
\begin{itemize}
\item Combined Fourier neural operator and neural ODE to provide a spatial-temporal continuous framework for modeling unknown dynamical systems based on data in various resolutions.
\item Demonstrated the long-term stable simulations of the trained models for three different dynamical systems with discontinuous features or chaotic behaviors.
\item Proposed a hybrid optimization scheme that efficiently leverages both the short-term time series and long-term statistics data.
\end{itemize}

\section{Methodology}
\subsection{Problem Setting}
\label{ssec:problem_setting}

We focus on the continuous dynamical system in a general form:
\begin{equation}
    \label{eq:CSCT}
    \frac{\partial \boldsymbol{u}(\boldsymbol{x},t)}{\partial t} = \mathcal{G}(\boldsymbol{u}(\boldsymbol{x},t), t),
\end{equation}
with spatial variable $\boldsymbol{x} \in D_{{\boldsymbol{x}}} \subseteq \mathbb{R}^{d_{\boldsymbol{x}}}$, temporal variable $t \in [0,T] \subset \mathbb{R}$, state function $\boldsymbol{u}(\cdot,\cdot) \in \mathcal{U}(D_{\boldsymbol{x}}\times [0,T];\mathbb{R}^{d_{\boldsymbol{u}}})$, spatial profile of state function at time $t$ is $\boldsymbol{u}(\cdot, t) \in \mathcal{U}_t(D_{\boldsymbol{x}}; \mathbb{R}^{d_{\boldsymbol{u}}})$, and $\mathcal{G}: \mathcal{U}_t \times [0,T] \mapsto \mathcal{U}_t$ is a non-linear operator. If the system is autonomous, i.e., the system does not depend on time, $\mathcal{G}$ would become a non-linear spatial operator. $\mathbb{R}^d$ is a real vector space with $d$ dimension, $D_{\boldsymbol{x}}$ is a bounded domain, $\mathcal{U}$ and $\mathcal{U}_t$ are separable Banach spaces of functions taking values in $\mathbb{R}^{d_{\boldsymbol{u}}}$. 

Assuming that the operator $\mathcal{G}$ in Eq.~\eqref{eq:CSCT} is unknown, we propose to learn a data-driven dynamical operator $\mathcal{\Tilde{G}}$ that approximates $\mathcal{G}$, based on the time series data of $\boldsymbol{u}$ and a combined use of neural ODE~\cite{chen2018neural} and neural operator~\cite{li2021fourier}. To enhance the performance of the learned model, the data-driven operator is trained to match both short-term trajectories and long-term statistics from the true system. A hybrid scheme of gradient-based and derivative-free methods is then demonstrated to facilitate efficient training with the combined types of data.

Given a time series $\{\boldsymbol{u}(t_n)\}_{n=0}^{N}$ of system state $\boldsymbol{u}$ in Eq.~\eqref{eq:CSCT}, where $\boldsymbol{u}(t):=\boldsymbol{u}(\cdot, t)|D_{\boldsymbol{x}}^{(M)} \in \mathbb{R}^{d_{\boldsymbol{u}}\times M}$ is discretization of $\boldsymbol{u}(\cdot,t)$ with $D_{\boldsymbol{x}}^{(M)}=\{x_1,...,x_M\} \subset D_{\boldsymbol{x}}$ is a $M$-point discretization of the domain $D_{\boldsymbol{x}}$, we aim to build a continuous spatial-temporal model for the system by constructing a parametric map $\mathcal{\Tilde{G}}$ with parameters $\boldsymbol{\theta}$ to approximate the non-linear operator $\mathcal{G}$. 
The continuous spatial-temporal model would then be:
\begin{equation}
    \label{eq:CSCT_Model}
    \frac{\partial \Tilde{\boldsymbol{u}}(\boldsymbol{x},t)}{\partial t} = \mathcal{\Tilde{G}}(\Tilde{\boldsymbol{u}}(\boldsymbol{x},t), t;\boldsymbol{\theta}).
\end{equation}
To match with the time series data $\{\boldsymbol{u}(t_n)\}_{n=0}^{N}$, the unknown parameters $\boldsymbol{\theta}$ in $\Tilde{\mathcal{G}}$ can be calibrated by minimizing the short-term loss for trajectory matching:
\begin{equation}
    \label{eq:opt}
    L_s := \sum_{n=0}^{N_s} J_s(\boldsymbol{u}(t_n),\ \Tilde{\boldsymbol{u}}(t_n)).
\end{equation}
where $\tilde{\boldsymbol{u}}(t_n) = \boldsymbol{u}_{t_0}+\int_{t_0}^{t_n}\Tilde{\mathcal{G}}( \boldsymbol{u}, t;\boldsymbol{\theta})\diff t$ is the predicted states, $\boldsymbol{u}(t_n)$ is the observed system states from the true system, and $J_s:\mathbb{R}^{d_{\boldsymbol{u}}\times M} \times \mathbb{R}^{d_{\boldsymbol{u}}\times M} \mapsto \mathbb{R}$ is a loss function at every time step. It is worth noting that the continuous counterpart of the total short-term loss $L_s$ is a loss functional operating on two space-time continuous functions and is defined by a squared norm for the Bochner space $L^2([0,T_s];L^2(D_{\boldsymbol{x}};\mathbb{R}^{d_{\boldsymbol{u}}}))$, where $T_s=t_{N_s}-t_0$ corresponds to the total time range of short-term trajectory matching. The continuous counterpart of the loss function $J_s$ is a functional on two spatial continuous functions which defined as $L^2(L^2(D_{\boldsymbol{x}};\mathbb{R}^{d_{\boldsymbol{u}}})\times L^2(D_{\boldsymbol{x}};\mathbb{R}^{d_{\boldsymbol{u}}}); \mathbb{R})$. $N_s$ is a hyper-parameter controlling the short-term state prediction horizon during training process. Given a long time series of states as data of a dynamical system, we sample a batch of short time series with length $N_s$ from the dataset, with which we can use mini-batch training to obtain the parameters $\boldsymbol{\theta}$.

To capture a long-term statistical property of the true dynamical system, the unknown parameters $\boldsymbol{\theta}$ in $\tilde{\mathcal{G}}$ can be calibrated by minimizing the long-term loss for statistics matching:
\begin{equation}
    \label{eq:opt_long}
    L_l\left(\beta\left(\{\boldsymbol{u}\left(t_n\right)\}_{n=0}^{N_l}\right), \ \beta\left(\{\tilde{\boldsymbol{u}}\left(t_n\right)\}_{n=0}^{N_l}\right)\right).
\end{equation} 
where $\beta: \mathbb{R}^{d_{\boldsymbol{u}} \times M \times N_l} \mapsto \mathbb{R}^{d_{\beta}}$ maps a long-term trajectory to its important statistics, $L_l: \mathbb{R}^{d_{\beta}} \times \mathbb{R}^{d_{\beta}} \mapsto \mathbb{R}$ is a loss function to quantify the differences of long-term statistics between the modeled system and the true one, and $N_l$ is another hyper-parameter that control the horizon over which the long-term statistics is calculated. The continuous counterpart of the map $\beta$ operates on a space-time continuous function in the space $C([0,T_l];L^2(D_{\boldsymbol{x}};\mathbb{R}^{d_{\boldsymbol{u}}}))$, where $T_l=t_{N_l}-t_0$ corresponds to the total time range of long-term statistics matching. In practice, to reduce the computational costs during the training process, we sample a batch of relatively long time series with length $N_l$ instead of calculating the long-term statistics for the entire time series dataset. The long-term horizon is typically much larger than the short-term horizon, i.e., $T_l \gg T_s$.

\subsection{Neural Operator}
\label{ssec:NeuralOp}
To construct the data-driven dynamical operator $\Tilde{\mathcal{G}}$ in the continuous setting, i.e., viewing the system state $\boldsymbol{u}$ as a continuous spatial-temporal field instead of a discrete finite-dimensional vector, we rely on the recent developments of neural operators~\cite{li2021fourier,lu2021learning}. As a high-level framework for modeling continuous dynamical systems, the neural dynamical operator can utilize any neural operator architecture to construct the data-driven dynamical operator $\tilde{\mathcal{G}}$. In this work, we present results using Fourier neural operator (FNO) ~\cite{li2021fourier}. In general, neural operators can approximate $\mathcal{G}$ in Eq.~\eqref{eq:CSCT} which is a non-linear mapping between infinite-dimensional spaces:
\begin{equation}
\label{eq:NO}
    \mathcal{G}: \mathcal{U}_t \times [0, T] \mapsto \mathcal{U}_t,
\end{equation}
with a neural network $\tilde{\mathcal{G}}(\cdot; \boldsymbol{\theta})$ parameterized by $\boldsymbol{\theta}$. The key advantages of continuous mapping in Eq.~\eqref{eq:NO} are: (i) the performance of the trained mapping is resolution-invariant, and (ii) the flexibility of using data with different discretizations.
The resolution-invariant feature allows the model to learn from and evaluate the functions discretized in an arbitrary way. In other words, the model trained at one specific resolution can be used to perform inference at different resolutions of input and output functions, without a significantly increased error level. With the resolution-invariant property in the construction of the FNO architecture, the training process can also potentially combine the use of data in different resolutions.

We then focus on the use of FNO to construct the data-driven operator $\tilde{\mathcal{G}}$. The FNO applies a Fourier transform to convert the input spatial function from the spatial domain to the Fourier domain, processes the data within the Fourier domain, and subsequently performs an inverse Fourier transform to transform the results back to the spatial domain. Similarly, when the input function is defined on a temporal domain, it can still be transformed between the temporal domain and the Fourier domain. More specifically, FNO first linearly transforms the state function $\boldsymbol{u}(\cdot, t) \subset \mathcal{U}_t$ and temporal variable $t$ to lift the dimension from $\mathbb{R}^{d_{\boldsymbol{u}}+1}$ to $\mathbb{R}^{d_{\boldsymbol{v}}}$: $\boldsymbol{v}_0(\boldsymbol{x}) = P(\boldsymbol{u}(\boldsymbol{x}, t), t)$. Then $\boldsymbol{v}_0(\boldsymbol{x})$ will be iteratively transformed by Fourier layers with the output dimension staying the same as $d_{\boldsymbol{v}}$:
\begin{equation}
    \label{eq:FNO}
    \boldsymbol{v}_{n+1}(\boldsymbol{x}) = \sigma(\boldsymbol{W}\boldsymbol{v}_n(\boldsymbol{x}) + (\boldsymbol{K}\boldsymbol{v}_n)(\boldsymbol{x})),\quad n=0,1 \ldots N-1,
\end{equation}
where $(\boldsymbol{K}\boldsymbol{v}_n)(\boldsymbol{x}) = \mathcal{F}^{-1}(\boldsymbol{R} \cdot (\mathcal{F}\boldsymbol{v}_n))(\boldsymbol{x})$ is the Fourier integral operator, $\mathcal{F}$ is Fourier transform and $\mathcal{F}^{-1}$ is its inverse, $\boldsymbol{R}$ is the complex-valued parameters in the neural network, $\boldsymbol{W}$ is a linear transformation, and $\sigma$ is a nonlinear activation function. At last,  $\boldsymbol{v}_N(\boldsymbol{x})$ will be linearly transformed again to ensure the dimension of the final output same as the dynamics function in the space $\mathcal{U}_t$, i.e., $Q(\boldsymbol{v}_N(\boldsymbol{x})) = \frac{\partial \tilde{\boldsymbol{u}}(\boldsymbol{x}, t)}{\partial t} \in \mathbb{R}^{d_{\boldsymbol{u}}}$.

The classical loss function for training FNO quantifies the differences between the true dynamics $\frac{\partial \boldsymbol{u}(\boldsymbol{x},t)}{\partial t}$ and the predicted dynamics $\frac{\partial \tilde{\boldsymbol{u}}(\boldsymbol{x}, t)}{ \partial t}$. However, the data of $\frac{\partial \boldsymbol{u}(\boldsymbol{x},t)}{\partial t}$ may not be available in many applications, e.g., when the temporal resolution in the data of $\boldsymbol{u}(\boldsymbol{x},t)$ is not fine enough to obtain an informative estimation of its time derivative. Therefore, we may not be able to directly train the neural dynamical operator $\tilde{\mathcal{G}}$ by solving the optimization problem as typically done in the classical usage of FNO. To address this issue, we explore the training of a neural dynamical operator via the framework of neural ODE~\cite{chen2018neural}.

\subsection{Neural ODE}
\label{ssec:NeuralODE}

The goal of neural ODE is to train the surrogate model $\tilde{\mathcal{G}}$, which is constructed via FNO in this work, to approximate $\mathcal{G}$ in Eq.~\eqref{eq:CSCT} based on time series data $\{ \boldsymbol{u}(t_n) \}_{n=0}^N$ with $\boldsymbol{u}(t):=\boldsymbol{u}(\cdot, t)|D_{\boldsymbol{x}}^{(M)}$ which is defined in Section~\ref{ssec:problem_setting}. With a given $\tilde{\mathcal{G}}$ and initial state of true system $\boldsymbol{u}(t_0)$, the predictive system state at future time $t_n$ can be written as $\tilde{\boldsymbol{u}}(t_{n}) = \boldsymbol{u}(t_0) + \int_{t_0}^{t_{n}} \tilde{\mathcal{G}}(\tilde{\boldsymbol{u}}(t), t; \boldsymbol{\theta}) \diff t$, which can be obtained by an ODE solver in real applications. The trainable parameters $\boldsymbol{\theta}$ in $\tilde{\mathcal{G}}$ can then be obtained by minimizing the loss function in Eq.~\eqref{eq:opt}. With the loss function $J_s$ at each time step chosen as standard $\ell^2$-norm in this work, the total short-term loss for trajectory matching becomes 
\begin{equation}
\label{eq:opt2}
    L_s = \sum_{n=0}^{N_s} \|\boldsymbol{u}(t_n)- \tilde{\boldsymbol{u}}(t_n)\|^2.
\end{equation}

The neural ODE framework~\cite{chen2018neural} discussed two methods to calculate $\diff L_s / \diff \boldsymbol{\theta}$, including (i) backpropagation and (ii) adjoint sensitivity method, with which we can determine the optimal $\boldsymbol{\theta}$ by minimizing the short-term loss in Eq.~\eqref{eq:opt2} via gradient descent.

Backpropagation is a classical gradient descent optimization method for training a neural network. One drawback of the backpropagation for neural ODE training is the memory cost to store the results of its forward pass, which is linearly proportional to the number of $\tilde{\mathcal{G}}$ evaluations. Although various types of memory management techniques~\cite{margossian2019review} have been developed for automatic differentiation, the memory cost for matching a long time series data with a large scale model of $\tilde{\mathcal{G}}$ can still be inefficient or even infeasible.

On the other hand, the adjoint sensitivity method calculates the gradient $\diff L_s / \diff \boldsymbol{\theta}$ by a forward and a backward ODE integration. The forward integration solves for the state $\tilde{\boldsymbol{u}}$ at future time step $t_1,t_2,...,t_N$, with the initial state $\tilde{\boldsymbol{u}}(t_0)=\boldsymbol{u}(t_0)$. By introducing an adjoint state $\boldsymbol{a}(t_n)=\frac{\partial L_s}{\partial \tilde{\boldsymbol{u}}(t_n)}$, the backward integration solves for an augmented states $[\tilde{\boldsymbol{u}}(t), \boldsymbol{a}(t), \frac{\partial L_s}{\partial \boldsymbol{\theta}}]^\top$ from $t_n$ to $t_{n-1}$, for $n=N,N-1,...,1$, with the initial state $\tilde{\boldsymbol{u}}(t_N)$ evaluated from the forward integration, $\boldsymbol{a}(t_N) = \frac{\partial L_s}{\partial \tilde{\boldsymbol{u}}(t_N)}$, and $\frac{\partial L_s}{\partial \boldsymbol{\theta}}(t_N) = \boldsymbol{0}$:
\begin{equation}
\begin{aligned}
&\frac{\diff \tilde{\boldsymbol{u}}(t)}{\diff t} = \tilde{\mathcal{G}}(\tilde{\boldsymbol{u}}(t), t; \boldsymbol{\theta}), \\
&\frac{\diff \boldsymbol{a}(t)}{\diff t} = -\boldsymbol{a}(t)^\top\frac{\partial \tilde{\mathcal{G}}(\tilde{\boldsymbol{u}}(t), t; \boldsymbol{\theta})}{\partial \tilde{\boldsymbol{u}}(t)}, \\
&\frac{\diff}{\diff t}\frac{\partial L_s}{\partial \theta} = -\boldsymbol{a}(t)^\top\frac{\partial \tilde{\mathcal{G}}(\Tilde{\boldsymbol{u}}(t),t; \boldsymbol{\theta})}{\partial \boldsymbol{\theta}}.
\end{aligned}
\end{equation}

It should be noted that the solved adjoint state $\boldsymbol{a}(t_n)$ needs to be adjusted at $n=N-1,N-2,...,0$ during the backward integration, by adding a term of $\frac{\partial L_s}{\partial \tilde{\boldsymbol{u}}(t_n)}$ to account for the fact that the loss function $L_s$ explicitly depends on the system state $\tilde{\boldsymbol{u}}(t_n)$. During the whole backward integration, the vector-Jacobian products $\boldsymbol{a}^\top \frac{\partial \tilde{\mathcal{G}}} {\partial \tilde{\boldsymbol{u}}}$ and $\boldsymbol{a}^\top \frac{\partial \tilde{\mathcal{G}}}{\partial \boldsymbol{\theta}}$ can be evaluated on the fly without storing them in the memory. Therefore, the adjoint sensitivity method has a constant memory usage $O(1)$ with respect to the integration time steps, i.e., the number of $\tilde{\mathcal{G}}$ evaluations.

For a short-term integration, the backpropagation often has no memory cost issue and is computationally faster than the adjoint method, considering that the Jacobian is stored and does not need to be evaluated on the fly in the backward pass. However, the memory cost issue would prevent the use of backpropagation if the data involves long-term integration, e.g., time-averaged statistics. In addition, gradient-based optimization can potentially encounter numerical issues (e.g., gradient blow-up) for the long-term information of chaotic systems. Therefore, backpropagation is used in this paper for short-term trajectory matching, while the long-term statistics of a chaotic dynamical system are incorporated by a derivative-free Kalman method, instead of the adjoint sensitivity method. More details about the derivative-free Kalman method can be found in Section~\ref{ssec:EKI}.

\subsection{Ensemble Kalman Inversion}
\label{ssec:EKI}

The neural dynamical operator $\tilde{\mathcal{G}}$ in Eq.~\eqref{eq:CSCT_Model} can be calibrated to capture the long-term statistics of the true dynamical system by solving the optimization problem in Eq.~\eqref{eq:opt_long} with ensemble Kalman inversion (EKI). Originated from ensemble Kalman filter~\cite{evensen1994sequential, evensen2003ensemble, evensen2009data}, EKI~\cite{iglesias2013ensemble,kovachki2019ensemble} has been developed as an optimization method to solve inverse problems. More specifically, the goal of EKI is to estimate the unknown parameters from noisy data in the general form of an inverse problem:

\begin{equation}
    \boldsymbol{y} = G(\boldsymbol{\theta}) + \boldsymbol{\eta},
    \label{eq:EKI_model}
\end{equation}
where $\boldsymbol{y}$ is a vector of observation data, $G$ denotes a forward map, $\boldsymbol{\theta}$ corresponds to the trainable parameters, $\boldsymbol{\eta}\sim \mathcal{N}(\boldsymbol{0}, \boldsymbol{\Sigma}_{\boldsymbol{\eta}})$ is often chosen as Gaussian random noises with a covariance matrix $\boldsymbol{\Sigma}_{\boldsymbol{\eta}}$. In this work, $\boldsymbol{y} = \beta(\{\boldsymbol{u}(t_n)\}_{n=0}^{N_l}) \in \mathbb{R}^{d_{\beta}}$ is the observed long-term statistics of the dynamical system, and the covariance matrix $\boldsymbol{\Sigma_{\boldsymbol{\eta}}} \in \mathbb{R}^{d_{\beta} \times d_{\beta}}$ is empirically chosen. In practice, the covariance matrix can also be determined by observing an ensemble of long-term statistics from the true system, with each ensemble member having a different initial condition. The forward map $G$ is a composition of data-driven dynamical operator $\tilde{\mathcal{G}}$, a time integral operator $\int$ that is numerically evaluated by an ODE solver, and $\beta$ that calculates the time-averaged statistics from a time series of system states, i.e., 

\begin{equation}
\begin{gathered}
\label{eq:EKI_G}
G(\boldsymbol{\theta}) := \beta(\{\tilde{\boldsymbol{u}}(t_n)\}_{n=0}^{N_l}) = \beta\Bigl(\bigl\{\boldsymbol{u}(t_0)+\int_{t_0}^{t_n} \tilde{\mathcal{G}}(\tilde{\boldsymbol{u}}(\boldsymbol{x},t),t; \boldsymbol{\theta})\diff t\bigr\}_{n=0}^{N_l}\Bigr).
\end{gathered}
\end{equation}

To match the observed data $\boldsymbol{y}$ and the output of forward map $G$, EKI is employed to identify the optimal parameters $\boldsymbol{\theta}$. The optimization problem in Eq.~\eqref{eq:opt_long} solved by EKI can be written as
\begin{equation}
\label{eq:opt_long2}
    \min_{\boldsymbol{\theta}} ||\boldsymbol{\Sigma}_{\boldsymbol{\eta}}^{-\frac{1}{2}} (\boldsymbol{y}-G(\boldsymbol{\theta}))||^2.
\end{equation}
To solve the optimization problem in Eq.~\eqref{eq:opt_long2}, the EKI formula that iteratively updates the ensemble estimation of parameters $\{\boldsymbol{\theta}^{(j)}\}_{j=1}^J$ is:
\begin{equation}
\begin{gathered}
    \label{eq:EKI_update}
    \boldsymbol{\theta}_{n+1}^{(j)} = \boldsymbol{\theta}_n^{(j)} + \boldsymbol{\Sigma}_{n}^{\boldsymbol{\theta} \boldsymbol{g}}(\boldsymbol{\Sigma}_{n}^{\boldsymbol{gg}} + \boldsymbol{\Sigma}_{\boldsymbol{\eta}})^{-1}(\boldsymbol{y}^{(j)} - \boldsymbol{g}_n^{(j)}),
\end{gathered}
\end{equation}
where the index $n$ denotes the $n$-th EKI step and $\boldsymbol{y}^{(j)}$ corresponds to the perturbed observation data $\boldsymbol{y}$ via sampling the noises $\boldsymbol{\eta}$. With the ensemble of parameters $\{\boldsymbol{\theta}_{n}^{(j)}\}_{j=1}^J$ at the $n$-th EKI step, the terms $\boldsymbol{g}_n^{(j)}$, $\boldsymbol{\Sigma}_{n}^{\boldsymbol{\theta g}}$, and $\boldsymbol{\Sigma}_{n}^{\boldsymbol{gg}}$ in Eq.~\eqref{eq:EKI_update} are calculated as:
\begin{equation}
\begin{gathered}
    \label{eq:EKI_ensemble}
    \bar{\boldsymbol{\theta}}_n = \frac{1}{J}\sum_{j=1}^J\boldsymbol{\theta}_n^{(j)}, ~~ \boldsymbol{g}^{(j)}_{n} = G(\boldsymbol{\theta}_n^{(j)}), ~~ \bar{\boldsymbol{g}}_n = \frac{1}{J}\sum_{j=1}^J\boldsymbol{g}_n^{(j)}, \\
    \boldsymbol{\Sigma}_{n}^{\boldsymbol{\theta} \boldsymbol{g}} = \frac{1}{J-1}\sum_{j=1}^J(\boldsymbol{\theta}_n^{(j)} - \bar{\boldsymbol{\theta}}_n)(\boldsymbol{g}_{n}^{(j)} - \bar{\boldsymbol{g}}_{n})^{\top}, \\
    \boldsymbol{\Sigma}_{n}^{\boldsymbol{gg}} = \frac{1}{J-1}\sum_{j=1}^J(\boldsymbol{g}_{n}^{(j)} - \bar{\boldsymbol{g}}_{n})(\boldsymbol{g}_{n}^{(j)} - \bar{\boldsymbol{g}}_{n})^{\top}.
\end{gathered}
\end{equation}

\subsection{Neural Dynamical Operator}
\label{ssec:NDO}

Based on the techniques introduced in Sections~\ref{ssec:NeuralOp} and \ref{ssec:NeuralODE}, we present a spatial-temporal continuous framework that learns a data-driven dynamical operator $\Tilde{\mathcal{G}}$ in Eq.~\eqref{eq:CSCT_Model} based on the short-term time series of the true system states. More specifically, the key components of the proposed framework include:
\begin{itemize}
\item Constructing the dynamical operator $\Tilde{\mathcal{G}}$ via a Fourier neural operator.
\item With the short-term time series of the true system states, updating the parameters $\theta$ of the dynamical operator $\Tilde{\mathcal{G}}$ via solving the optimization problem in Eq.~\eqref{eq:opt} based on a gradient-based optimization method (e.g., neural ODE).
\end{itemize}

The merits of learning a dynamical operator $\Tilde{\mathcal{G}}$ in Eq.~\eqref{eq:CSCT_Model} are:
\begin{itemize}
\item The flexibility of using non-uniform data points in both space and time.
\item The resolution-invariance of the trained model in both space and time.
\end{itemize}

Besides being good at predicting short-term system states, the trained modeled system in Eq.~\eqref{eq:CSCT_Model} also demonstrates a stable long-term simulation. There are some existing methods to make stable long-term predictions for chaotic dynamical systems such as reduced order modeling~\cite{maulik2020time} and stabilized neural ODE~\cite{linot2023stabilized}. Instead of preventing the high-wavenumbers amplification by dimension reduction or employing a linear damping term, the use of the Fourier neural operator in this work facilitates the stable long-term simulation by filtering out the high-order modes in the Fourier space. Therefore, neural dynamical operator trained via short-term time series data can still avoid a numerical explosion in long-term simulations and also qualitatively retain the statistical property of the true system.

However, the use of short-term time series data alone could not guarantee a quantitative match of the long-term statistics between the trained model and the true system. This limitation motivates the combined use of short-term time series and long-term time-averaged statistics as two types of data for training the neural dynamical operator. A hybrid optimization scheme that can efficiently incorporate both types of data is introduced in Section~\ref{ssec:derivative_hybrid}.

\begin{figure}[H]
    \centering
    \includegraphics[width=0.8\textwidth]{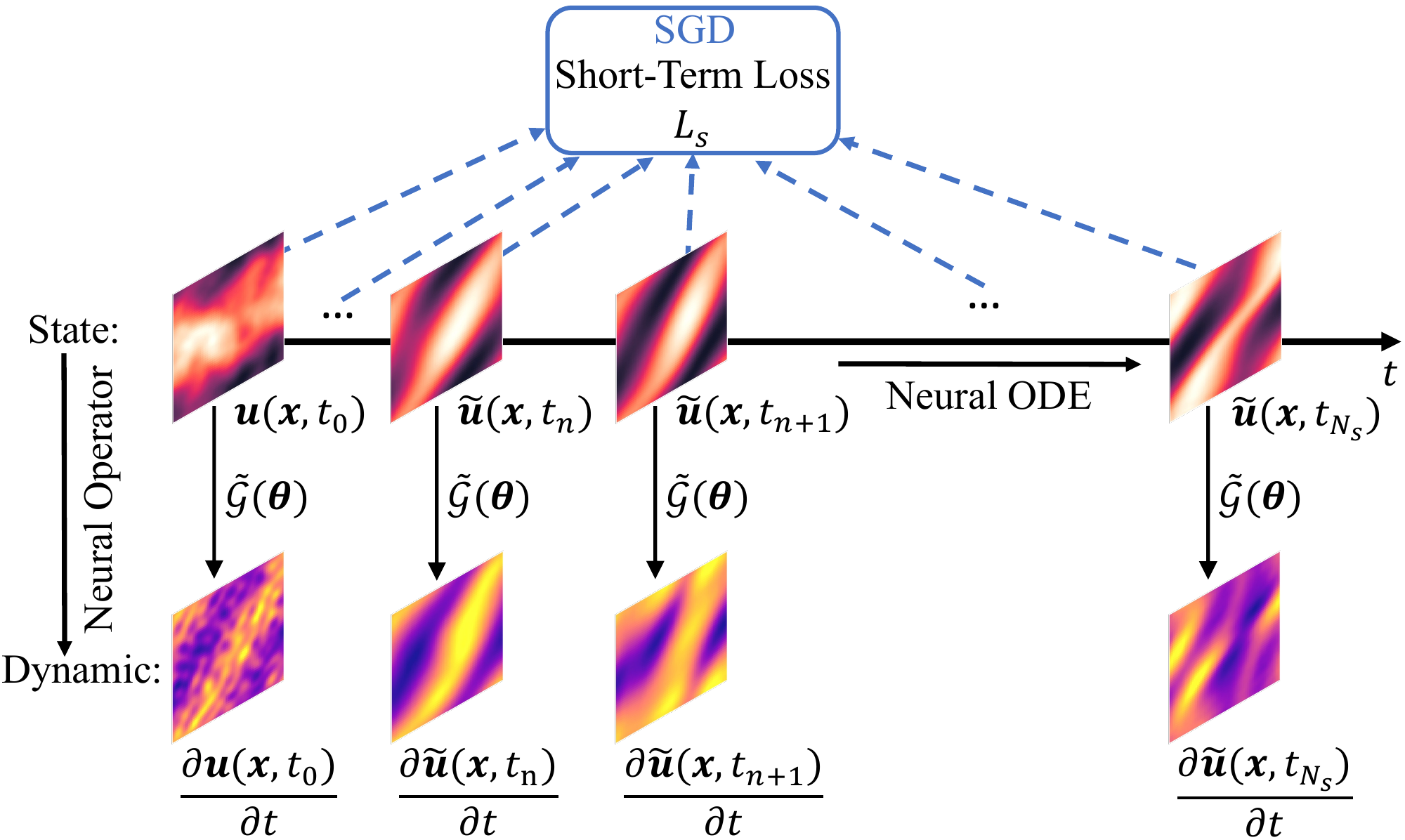}
    \caption{Schematic diagram of neural dynamical operator (based on Navier--Stokes equations). The dynamics of the system for the current state are approximated by neural operator, then the future states are evaluated with an ODE solver along with time given any initial state. The neural dynamical operator $\Tilde{\mathcal{G}}$ is trained by minimizing the Loss $L_s$ with gradient-based optimization.}
    \label{fig:SchematicDiagram1}
\end{figure}

\subsection{A Hybrid Optimization Scheme}
\label{ssec:derivative_hybrid}
To efficiently incorporate both short-term time series and long-term statistics of the true system states as the training data, we propose a hybrid optimization scheme that iteratively solves the optimization problems in Eqs.~\eqref{eq:opt} and~\eqref{eq:opt_long}. Within each iteration, the parameters $\boldsymbol{\theta}$ of the dynamical operator $\Tilde{\mathcal{G}}$ in Eq.~\eqref{eq:CSCT_Model} are first updated via a gradient-based optimization method (e.g., backpropagation) that solves Eq.~\eqref{eq:opt} with the short-term time series data of the true system states and then further adjusted via a derivative-free optimization method (e.g., ensemble Kalman inversion) to account for the long-term statistics data in Eq.~\eqref{eq:opt_long}. Based on the history of both short-term state error and long-term statistics error during the EKI updating, we select a robust $\boldsymbol{\theta}$ that yields relatively small values for both types of errors. From a more general perspective, the hybrid optimization aims to solve a dual-objective optimization problem: it seeks to minimize both the loss from short-term state prediction and the loss from long-term statistics matching. The optimization procedure leads to a set of candidate solutions that approximates the Pareto front, from which we can robustly select the solution that achieves a balanced performance of predicting the short-term trajectory and long-term statistics. The Algorithm~\ref{alg:HybridOptimization(Brief)} briefly summarizes the hybrid optimization scheme, while a detailed version is presented in Algorithm~\ref{alg:HybridOptimization} in \ref{sec:algorithm}.

\begin{algorithm}[H]
\caption{Training neural dynamical operator with the hybrid optimization scheme (A brief version)}
\label{alg:HybridOptimization(Brief)}
\begin{algorithmic}[1]
\For{$i \gets 1$ \textbf{to} $N_{\textrm{ep}}$}
        \State $\boldsymbol{\theta} \gets \boldsymbol{\theta} - \lambda \cdot \nabla_{\boldsymbol{\theta}} L_s$ \Comment{SGD Updating}
    \If { $i \bmod k=0 $ } 
        \For {$m \gets 1$ \textbf{to} $N_{\textrm{it}}$}
            \State $\{ \boldsymbol{\theta}^{(j)} \}_{j=1}^J \gets \{\boldsymbol{\theta}^{(j)} + \boldsymbol{\Sigma}^{\boldsymbol{\theta} \boldsymbol{g}}(\boldsymbol{\Sigma}^{\boldsymbol{gg}} + \boldsymbol{\Sigma}_{\boldsymbol{\eta}})^{-1}(\boldsymbol{y}^{(j)} - \boldsymbol{g}^{(j)}) \}_{j=1}^J$ \Comment{EKI Updating}
            \EndFor
        \State $\boldsymbol{\theta} \gets \frac{1}{J}\Sigma_{j=1}^{J} \boldsymbol{\theta}^{(j)}$
    \EndIf
\EndFor
\State \text{Select robust $\boldsymbol{\theta}^{\star}$ based on the error history during EKI updating.}
\end{algorithmic}
\end{algorithm}

The key merit of the hybrid optimization scheme is the efficient incorporation of both short-term time series and long-term time-averaged statistics data. With the use of both types of data, the trained dynamical operator is expected to have better generalization capability, which is confirmed by the numerical example of the Kuramoto–Sivashinsky equation in Section~\ref{ssec:KSE}. Though the two optimization processes are independent of each other, the selection of the covariance matrix $\boldsymbol{\Sigma}_{\boldsymbol{\eta}}$ in Eq.~\eqref{eq:opt_long2} will still affect results of hybrid optimization. The reason is that the same parameters in the neural dynamical operator are updated by both optimization process. If the covariance matrix $\boldsymbol{\Sigma}_{\boldsymbol{\eta}}$ is of a small magnitude, i.e., the observation noises of long-term statistics are small, the hybrid optimization would prefer smaller mismatches in long-term statistics from the true system, which could negatively impact the performance of the hybrid optimization, especially when the finite time-averaged approximation of the long-term statistics have larger intrinsic uncertainties compared to the level of observation noises in EKI. The intrinsic uncertainties due to the finite time-averaged approximation of the long-term statistics can be estimated by obtaining an ensemble observation of long-term statistics, which can be used to set up the observation noises of long-term statistics in EKI. In practice, a simpler choice for $\boldsymbol{\Sigma}_{\boldsymbol{\eta}}$ is a diagonal matrix, e.g., $c^2 \boldsymbol{\mathrm{I}}_{d_{\boldsymbol{y}}}$ where $c \in \mathbb{R}$ is a constant scalar and $\boldsymbol{\mathrm{I}}_{d_{\boldsymbol{y}}} \in \mathbb{R}^{d_{\boldsymbol{y}}\times d_{\boldsymbol{y}}}$ is an identity matrix with dimension $d_{\boldsymbol{y}}$. With this choice, the scalar $c$ plays the role of the hyper-parameter that controls the results of hybrid optimization. In this work, the constant $c$ is empirically chosen as $0.1$, based on the magnitude of intrinsic uncertainties due to finite time-averaged approximation of long-term statistics estimated by ensemble simulations. Alternatively, the hyper-parameter $c$ can be determined by cross-validation, i.e., choosing the optimal $c$ that provides the best cross-validation results.

\begin{figure}[H]
    \centering
    \includegraphics[width=0.8\textwidth]{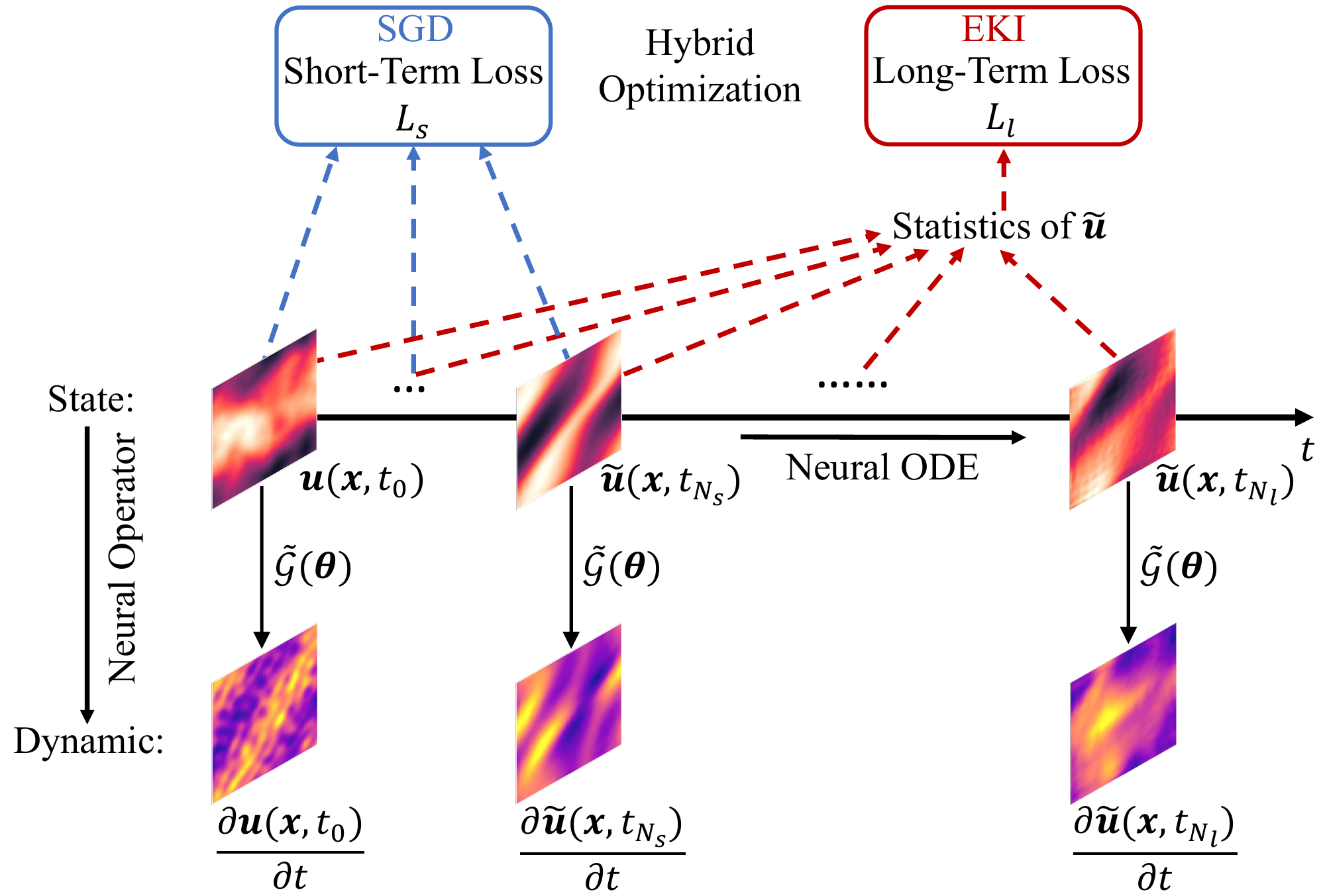}
    \caption{Schematic diagram of neural dynamical operator with hybrid optimization scheme (based on Navier--Stokes equations). To better generalize the model by utilizing both short-term and long-term data, the neural dynamical operator $\Tilde{\mathcal{G}}$ is trained by the hybrid optimization scheme which will iteratively update parameters by stochastic gradient descent (SGD) method to minimize short-term states loss $L_s$ and by derivative-free method (EKI) to minimize long-term statistics loss $L_l$. The short-term system evolution in $[t_0, t_{N_s}]$ corresponds to Fig.~\ref{fig:SchematicDiagram1}.}
    \label{fig:SchematicDiagram2}
\end{figure}

\section{Numerical Experiments}

We demonstrate the performance of the continuous spatial-temporal model on three examples, including (i) 1-D viscous Burgers' equation, (ii) 2-D Navier--Stokes equations, and (iii) Kuramoto--Sivashinsky equation. Short-term time series data generated from each true system are sub-sampled in both spatial and temporal domain with various resolutions, to confirm the resolution-invariance of the trained model with respect to both spatial and temporal discretizations. For all the examples, we also show the stable long-term simulation with the trained models, which is mainly due to the high-wavenumber filtering in the Fourier neural operator. For the example of Kuramoto--Sivashinshky equation, we present a quantitative comparison of the long-term statistics between the model trained with short-term time series data and the one trained with both short-term time series and long-term statistics data. The results demonstrate the merit of the combined use of data via the proposed hybrid optimization scheme. For all the examples, the fixed-step Runge-Kutta method (RK4) is used as the ODE solver, the absolute error is the mean squared error, and the relative error is calculated by the average of $\frac{||u-\Tilde{u}||_2}{||u||_2}$ across sample size where $||\cdot||_2$ is $\ell^2$-norm.

\subsection{Viscous Burgers' Equation}
    
Burgers' equation is a classical example of partial differential equation and has been widely used in many fields such as fluid mechanics~\cite{burgers1948mathematical, hopf1950partial, bec2007burgers}, nonlinear dynamics~\cite{kardar1986dynamic, weinan2000invariant} and traffic flow~\cite{helbing2001traffic, nagatani2002physics}. The governing equation of the viscous Burgers' equation is:

\begin{equation}
\begin{aligned}
\dfrac{\partial u}{\partial t} = -u\dfrac{\partial u}{\partial x} + \nu \dfrac{\partial^2 u}{\partial x^2},
\end{aligned}
\label{eq:VBE}
\end{equation}
where $x \in (0, L_x)$ with periodic boundary conditions, $t \in [0, L_t]$, $\nu$ is the viscosity coefficient, and $u(x,0)$ is a given initial condition. Neural dynamical operator aim to learn the operator on the right-hand-side of Eq.~\eqref{eq:VBE}, i.e., $\mathcal{G}: u \mapsto \frac{\partial u}{\partial t}$. By studying this example, we demonstrate that the trained neural dynamical operator has resolution-invariance with respect to both spatial and temporal discretizations. Also, the trained model can capture the shock behavior and shows a good performance in predicting the trajectory of the true system in the test dataset.

The simulation settings are $L_x = 1, L_t = 5, \nu=10^{-3}, dx = 1/1024, dt = 0.005$. We perform 1000 simulations as training data and another 100 simulations as test data, and the initial conditions of these simulations are randomly sampled from a 1-D Gaussian random field $\mathcal{N}(0, 625(-\Delta +25I)^{-2})$ with periodic boundary conditions, where $\Delta$ is a Laplace operator and $I$ is an identity operator. This choice of Gaussian random field is the same as in the paper of FNO~\cite{li2021fourier}.

In the optimization problem in Eq.~\eqref{eq:opt}, we first simulate the modeled system for $L_t=5$ time units given an initial state $u(t_0)$ to obtain the trajectory $\{\tilde{u}(t_n)\}_{n=1}^{N_s}$ with $t_{N_s} = 5$ and then minimize the $\ell^2$-norm between the simulated system trajectory and the true one. During the training process, we select a small subset from 1000 training simulations as a mini-batch to perform the gradient descent optimization in each epoch.

To construct the neural dynamical operator $\Tilde{\mathcal{G}}$, we use FNO as a surrogate model with $d_v=64$ and $k_{max}=24$. $d_v$ is a higher dimension where the input data be lifted to and $k_{max}$ is a cut point above which the modes of Fourier transform of input data will be truncated. We train the model with $10^3$ epochs with 10 simulations as one data batch in each epoch. The optimizer is Adam with $10^{-3}$ learning rate and cosine annealing schedule. 

We train the neural dynamical operator based on short-term time series data with various resolutions in both space and time. The test errors are summarized in Table \ref{tab:VBE_resolution}. To demonstrate the resolution-invariance of the trained models in both space and time, we present two types of test errors in Table~\ref{tab:VBE_resolution}: the Test Error (I) is based on the test data of the same resolution $dx=1/1024$ and $dt=0.05$, and the Test Error (II) is based on resolution setting same as each train data. By comparing these two types of test errors with a fixed training resolution, it can be seen that the test error stays at the same order of magnitude when predicting on a finer resolution test dataset. On the other hand, the comparison among the cases with different training data resolutions demonstrates that the trained model can still perform well with a relatively sparse temporal resolution. The resolution-invariance property of the trained neural dynamical operator makes it flexible in using training data with low or even mixed resolutions.

\begin{table}[H]
\begin{center}
\caption{The test errors of viscous Burgers' equation with various resolution settings for train data. The Test Error (I) is based on the test data of the resolution $dx=1/1024$ and $dt=0.05$, while the Test Error (II) is based on a resolution setting the same as each training data.}
\label{tab:VBE_resolution}

\begin{tabular}{|c|cc|cc|cc|}
\hline
\multicolumn{1}{|c|}{\multirow{2}{*}{\diagbox[width=8em]{Resolution}{Error}}}  & \multicolumn{2}{c|}{Train Data}  & \multicolumn{2}{c|}{Test Error (I)} & \multicolumn{2}{c|}{Test Error (II)}  \\ \cline{2-7} 

\multicolumn{1}{|c|}{} & \multicolumn{1}{c|}{$dx$} & $dt$  & \multicolumn{1}{c|}{Absolute} & \multicolumn{1}{c|}{Relative} & \multicolumn{1}{c|}{Absolute} & Relative \\ \hline

Resolution1 & \multicolumn{1}{c|}{1/512}  & 0.05 & \multicolumn{1}{c|}{6.7010e-05} & 4.4004e-02  & \multicolumn{1}{c|}{6.7595e-05} & 4.4163e-02  \\ \hline

Resolution2 & \multicolumn{1}{c|}{1/256}  & 0.1 & \multicolumn{1}{c|}{6.6947e-05} & 4.3981e-02 & \multicolumn{1}{c|}{6.8292e-05} & 4.4299e-02 \\ \hline

Resolution3 & \multicolumn{1}{c|}{1/64}  & 0.5 & \multicolumn{1}{c|}{7.7509e-05} & 4.7405e-02 & \multicolumn{1}{c|}{7.3959e-05} & 4.7423e-03 \\ \hline

\end{tabular}
\end{center}
\end{table}

\begin{figure}[H]
    \centering
    \includegraphics[width=1\textwidth]{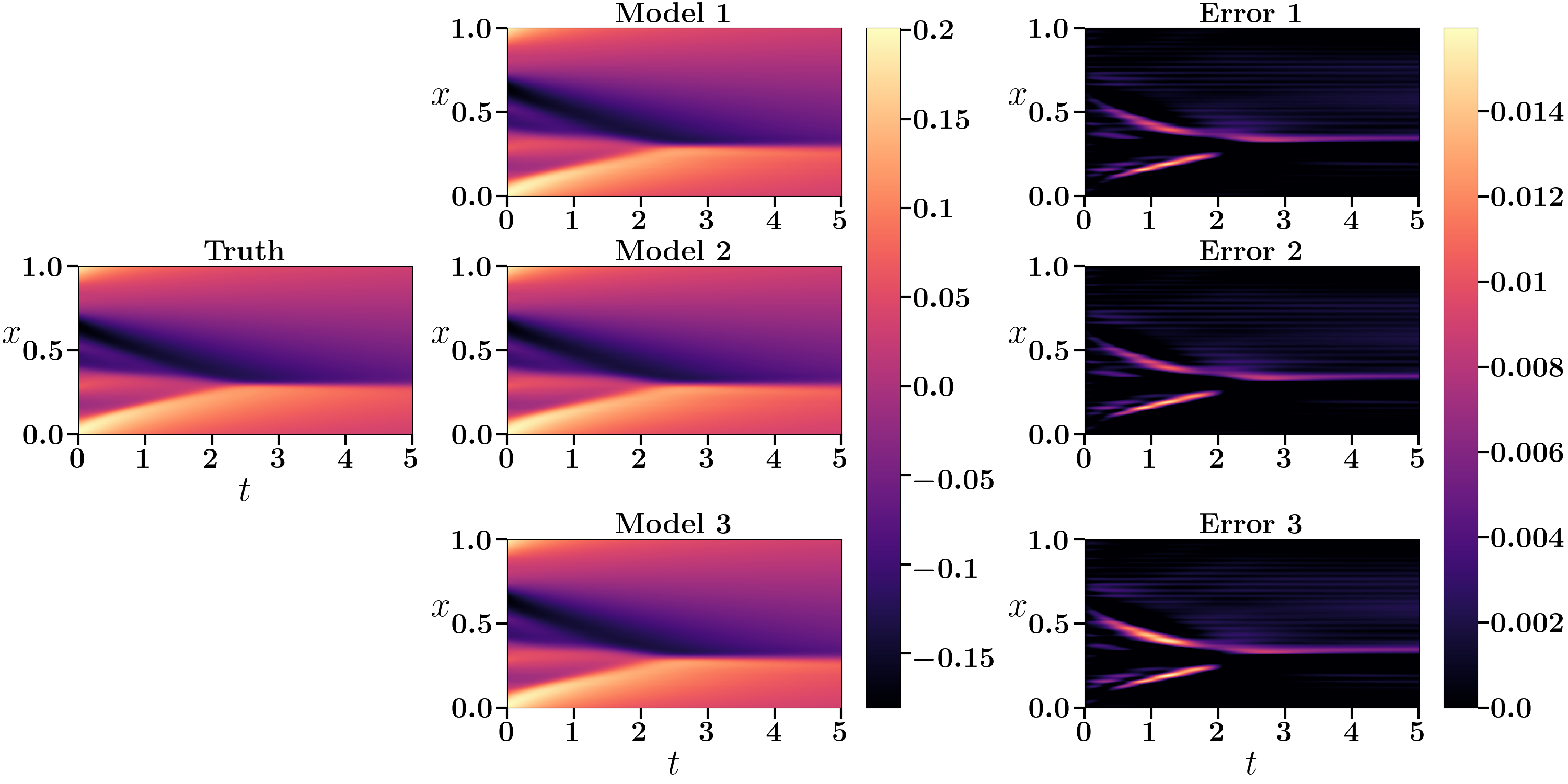}
    \caption{The spatial-temporal solutions of viscous Burgers' equation. Left column: true system. Middle column: trained models from three different resolutions with the same test data resolution ($dx=1/1024, dt=0.05$). Right column: errors of the solutions simulated based on the trained models. The index in the three trained models corresponds to the resolution settings in Table~\ref{tab:VBE_resolution}.}
    \label{fig:VBE_pcolor}
\end{figure}

\begin{figure}[H]
    \centering
    \includegraphics[width=1\textwidth]{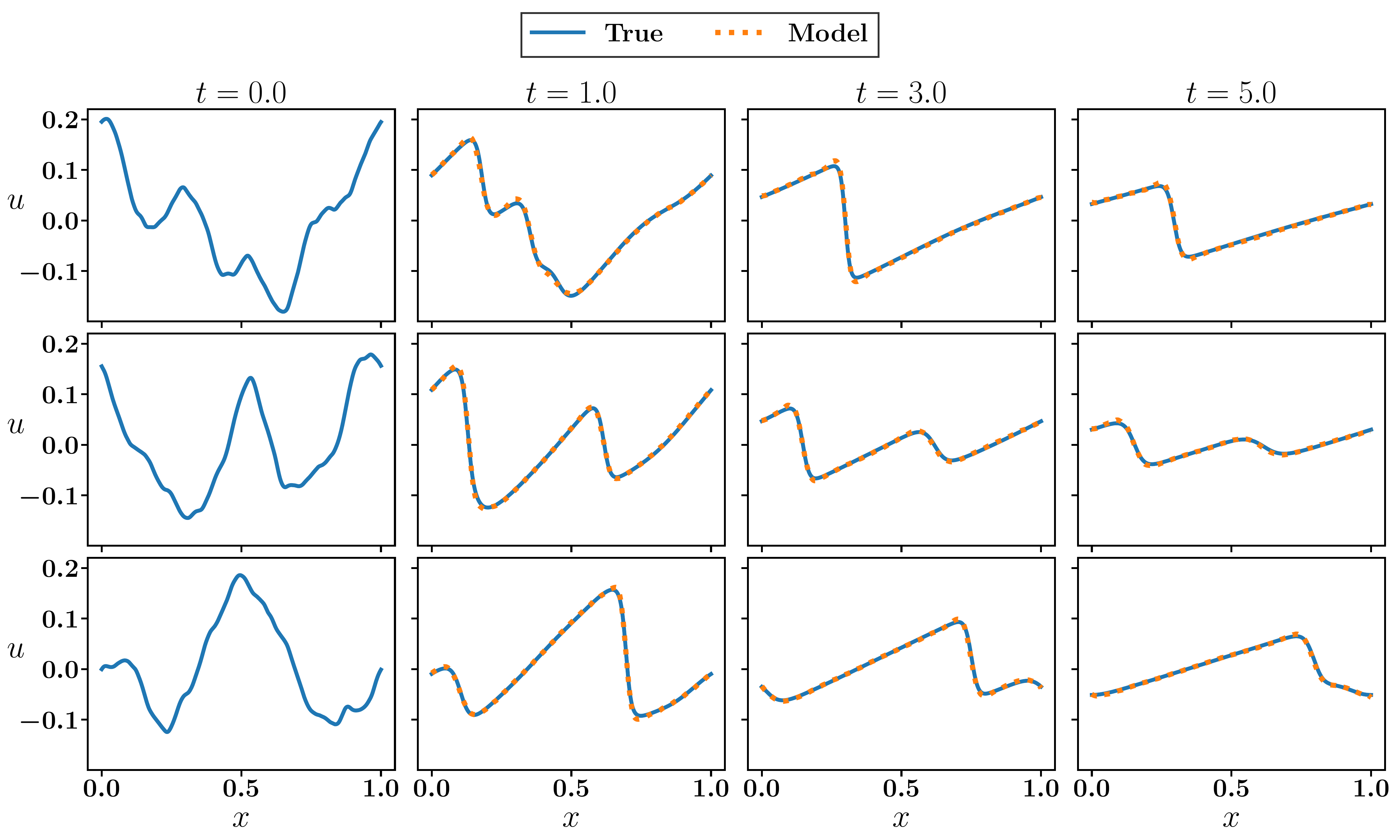}
    \caption{Solution profiles of viscous Burgers' equation for the true system and the model ones with different initial conditions from test data. The model is trained in a coarse resolution ($dx=1/64, dt=0.5$) and tested on a finer resolution ($dx=1/1024, dt=0.05$).}
    \label{fig:VBE_profile}
\end{figure}

The Burgers' equation can develop shocks over time in the absence of viscous term. In Eq.(\ref{eq:VBE}), The non-linear convective term $u\frac{\partial u}{\partial x}$ can result in a shock in the solution of $u$ (i.e., discontinuity of $u$ in space) when $\nu=0$. With $\nu \neq 0$, the diffusion term $\nu \frac{\partial^2 u}{\partial x^2}$ will lead to a continuous solution of $u$, while the spatial gradient of $u$ can be large at certain locations if $\nu$ is small. In this example, we choose a relatively small value of viscosity (i.e., $\nu=10^{-3}$ so that the solutions of the true system have such a feature of large spatial gradients. The results in Figs.~\ref{fig:VBE_pcolor} and~\ref{fig:VBE_profile} confirm that the trained models can capture this behavior and provide good performance for short-term trajectory prediction of the solutions of $u$.

In Fig.~\ref{fig:VBE_pcolor}, we present the spatial-temporal plots of the solutions from the true system and the modeled ones, with the initial condition sampled from test data. The left column presents the true solution of $u$. The middle column corresponds to the solutions from the modeled systems trained with different resolution settings in Table.~\ref{tab:VBE_resolution}, and we test the prediction performance of these trained models with the same resolution setting (i.e., $dx=1/1024$ and $dt=0.05$, which is finer than all the training resolutions). The left column shows the differences between the true solution and the solutions from the modeled systems. We can see that all predictions (shown in the middle column in Fig.~\ref{fig:VBE_pcolor}) capture the overall pattern of the true system. On the other hand, the absolute errors made by those predictions show that all trained models can provide relatively small errors at most locations and times, even testing on a finer resolution and starting with an unseen initial condition. It should be noted that the absolute errors tend to be larger close to the regions where the true solution has a large spatial gradient, but the magnitudes of those errors are still small compared to the true solution in those regions.

We also study the comparison of the energy spectrum between the true system and the model ones. The energy spectrum is defined as:
\begin{equation}
    \label{eq:ES}
    E(k,t) = \frac{1}{2}|u_f(k,t)|^2,
\end{equation}
where $u_f = \mathcal{F}u$ denotes the Fourier transform of the solution $u$ with $k$ as the wavenumber. Here $|\cdot|$ evaluates the magnitude of a complex number.

\begin{figure}[H]
    \centering    
    \includegraphics[width=1\textwidth]{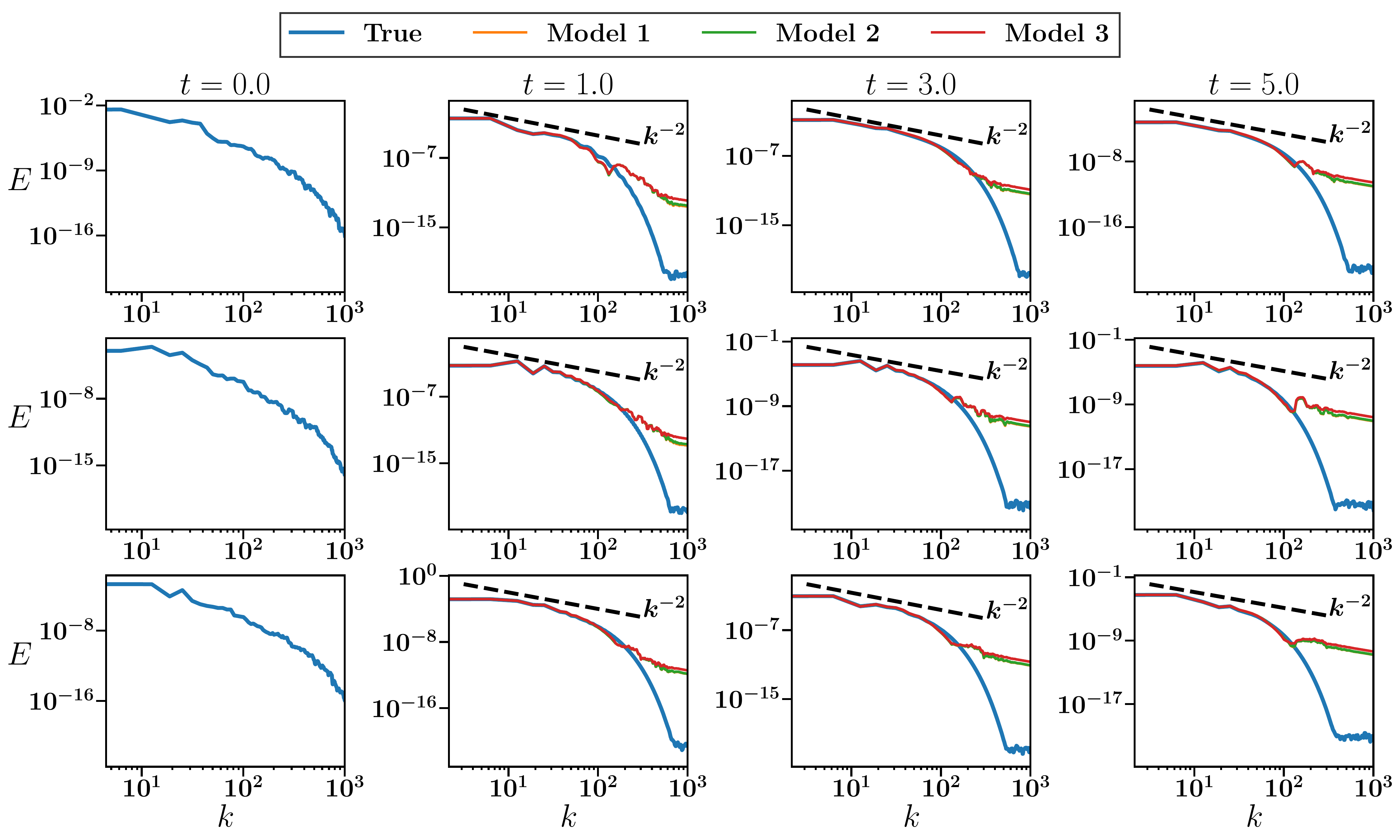}
    \caption{Energy spectrum of the true system and the model ones with different test initial conditions (in rows) and at different times (in columns). The index in the three trained models corresponds to the resolution settings in Table~\ref{tab:VBE_resolution}.}
    \label{fig:VBE_es}
\end{figure}

Fig.~\ref{fig:VBE_es} presents the energy spectrum of the solution $u$ for the true system and the model ones. The spectrum shows a slope $k^{-2}$, which agrees well with~\cite{linot2023stabilized}. The three trained models are tested with the same resolution ($dx=1/1024$, $dt=0.05$). It can be seen that the energy spectrum of the trained models has good agreement with each other, indicating that the resolution-invariance is achieved by the proposed neural dynamical operator. In addition, we can see that the trained models can capture the true spectrum up to a wavenumber of approximately $200$ to $300$. The main reason for the noticeable difference from the true system for larger wavenumbers is that all three initial conditions are not included in the training data. It is promising to see that the trained models still generalize quite well with all these unseen initial conditions in most wavenumbers and only start to display mismatches for very high wavenumbers. 

\subsection{Navier–Stokes Equations}

We consider the Navier–Stokes equations to study the performance of neural dynamical operator on a 2-D continuous dynamical system. The Navier--Stokes equations are partial differential equations that characterize the conservation of linear momentum in fluid flows~\cite{navier1822memoire, chorin1968numerical,acheson1991elementary,temam2001navier}. The 2-D Navier–Stokes equations written in the form of vorticity are:

\begin{equation}
\begin{gathered}
    \label{eq:NSE}
    \frac{\partial \omega}{\partial t} = -u\cdot \nabla \omega + \nu \Delta \omega + f, \\ 
    \nabla \cdot u = 0,
\end{gathered}
\end{equation}
where $u$ denotes a 2-D velocity vector, $\omega:=\nabla \times u$ represents the vorticity, $\nu$ is the kinematic viscosity of the fluid, and $f$ corresponds to a forcing function. Neural dynamical operator aim to learn the operator on the right-hand-side of Eq.~\eqref{eq:NSE}, i.e., $\mathcal{G}: \omega \mapsto \frac{\partial \omega}{\partial t}$. In real applications, the training data generated by simulations may not be with enough high resolution to well capture the true operator $\mathcal{G}$, e.g., high Reynolds number wall-bounded turbulent flows, for which resolving the Kolmogorov scales is still infeasible for many real engineering applications. With this example, we demonstrate that the trained neural dynamical operator can still capture the resolved information from a dataset, even with relatively low spatial resolution. However, the trained operator may not well characterize the true continuous operator if the training data is generated by simulations with too coarse spatial resolutions, and thus the prediction results on a higher spatial resolution could lead to larger errors.

The simulation domain is $\Omega=(0,1)^2$ with periodic boundary conditions in both $x$ and $y$ directions, and we simulate the system for the time $t \in [0,L_t]$ with $L_t=20$. The detailed settings are $dx=1/256, dy=1/256, dt=10^{-4}, \nu=10^{-3}$, and $f = 0.1(\sin(2\pi(x+y))+\cos(2\pi(x+y)))$. With initial condition $\omega(x,y,0)$ randomly sampled from a 2-D Gaussian random field $\mathcal{N}(0, 7^{1.5}(-\Delta + 49I)^{-2.5})$, we perform 1100 simulations in total, with 1000 simulations as training data and the other 100 simulations as test data.

To construct the neural dynamical operator $\mathcal{\Tilde{G}}$,  we use a 2-D FNO as surrogate model with $k_{\textrm{max},1}=12$, $k_{\textrm{max},2}=12, d_v=32$ for Resolution 1 to 4 and $k_{\textrm{max},1}=8$, $k_{\textrm{max},2}=8, d_v=24$ for Resolution 5 in Table.~\ref{tab:NSE_resolution}. The neural dynamical operator is trained with $2\times 10^4$ epochs, and one simulation from the training dataset serves as one mini-batch data (i.e., 1 training batch in each epoch). The optimizer is Adam with $10^{-3}$ learning rate and cosine annealing schedule.

We train the neural dynamical operator based on time series data with various spatial-temporal resolutions. The test errors are summarized in Table \ref{tab:NSE_resolution}: the Test Error (I) is based on the test data of the same resolution $dx=1/64, dy=1/64$ and $dt=0.2$, and the Test Error (II) is based on the resolution setting same as each train data. From the Test Error (II), it can be seen that models trained from all resolutions can achieve a small test error and stay at the same order of error magnitude when predicting on a test dataset whose resolution is the same as train data. However, from the Test error (I) in the rows of Resolution 4 and 5, we can see that models trained from a coarse resolution fail to show good performance when testing with higher data resolution. On the other hand, from the Test Error (I) in rows of Resolutions 1 to 3, we can still confirm that the temporal resolution-invariance property is achieved by the trained dynamical operator.

\begin{table}[H]
\begin{center}
\caption{The test errors of Navier--Stokes equation with various resolution settings for train data. The Test Error (I) is based on the test data of the resolution $dx=1/64$, $dy=1/64$, and $dt=0.2$, while the Test Error (II) is based on a resolution setting the same as each training data.}
\label{tab:NSE_resolution}

\begin{tabular}{|c|ccc|cc|cc|}
\hline
\multicolumn{1}{|c|}{\multirow{2}{*}{\diagbox[width=8em]{Resolution}{Error}}}  & \multicolumn{3}{c|}{Train Data}  & \multicolumn{2}{c|}{Test Error (I)} & \multicolumn{2}{c|}{Test Error (II)}  \\ \cline{2-8} 

\multicolumn{1}{|c|}{} & \multicolumn{1}{c|}{$dx$} & \multicolumn{1}{c|}{$dy$} & \multicolumn{1}{c|}{$dt$} & \multicolumn{1}{c|}{Absolute} & \multicolumn{1}{c|}{Relative} & \multicolumn{1}{c|}{Absolute} & Relative \\ \hline

Resolution1 & \multicolumn{1}{c|}{1/64} & \multicolumn{1}{c|}{1/64} & 0.2 & \multicolumn{1}{c|}{2.2011e-04} & 2.6937e-02 & \multicolumn{1}{c|}{2.2011e-04} & 2.6937e-02  \\ \hline

Resolution2 & \multicolumn{1}{c|}{1/64} & \multicolumn{1}{c|}{1/64} & 0.4 & \multicolumn{1}{c|}{2.1764e-04} & 2.6753e-02 & \multicolumn{1}{c|}{2.1739e-04 } & 2.6827e-02 \\ \hline

Resolution3 & \multicolumn{1}{c|}{1/64} & \multicolumn{1}{c|}{1/64} & 1 & \multicolumn{1}{c|}{2.0775e-04} & 2.6083e-02  & \multicolumn{1}{c|}{2.0591e-04} & 2.6089e-02 \\ \hline

Resolution4 & \multicolumn{1}{c|}{1/32} & \multicolumn{1}{c|}{1/32} & 0.2 & \multicolumn{1}{c|}{1.3277e-01} & 5.0054e-01 & \multicolumn{1}{c|}{2.2525e-04} &  2.7820e-02 \\ \hline

Resolution5 & \multicolumn{1}{c|}{1/16} & \multicolumn{1}{c|}{1/16} & 0.2 & \multicolumn{1}{c|}{3.5572e-01} & 7.9330e-01 & \multicolumn{1}{c|}{3.1081e-04} & 3.2806e-02 \\ \hline

\end{tabular}
\end{center}
\end{table}

\begin{figure}[H]
    \centering
    \includegraphics[width=0.9\textwidth]{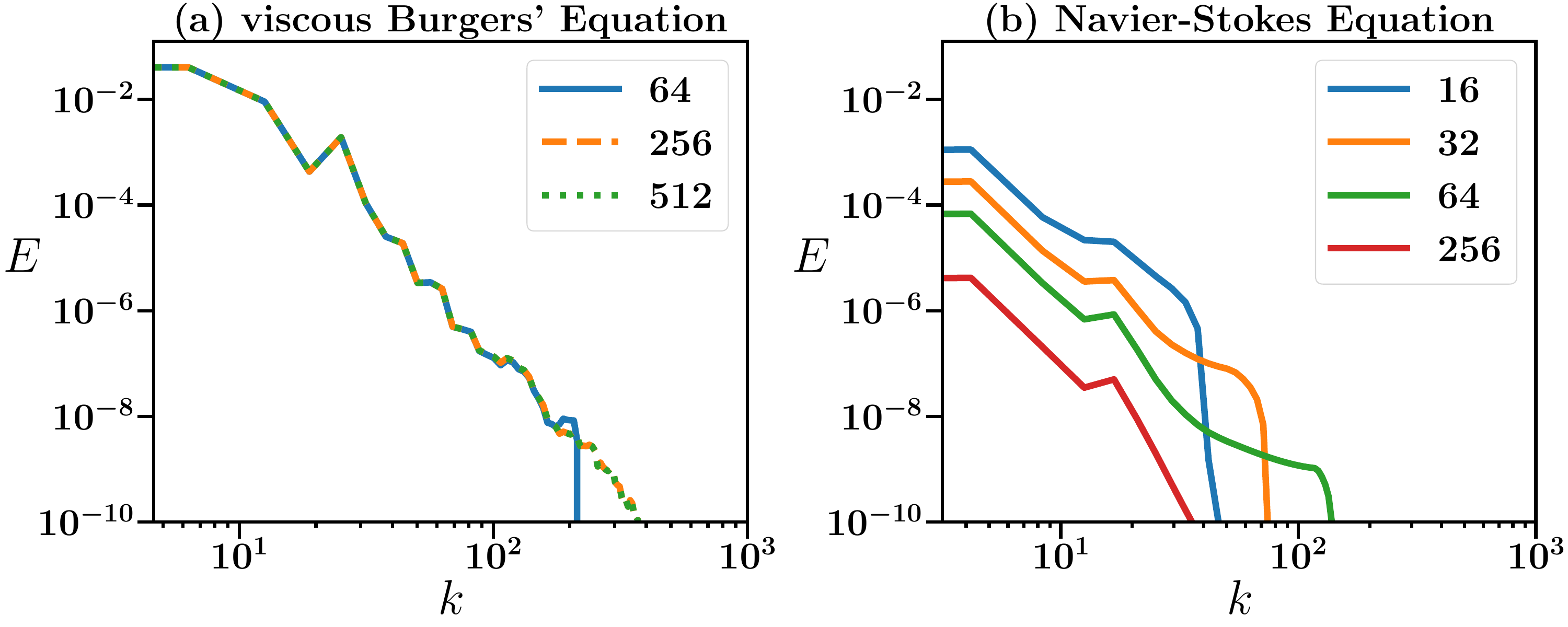}
    \caption{Energy spectrum of initial condition data of viscous Burgers' equation and Navier--Stokes equation with respect to different resolution settings in the Table~\ref{tab:VBE_resolution} and Table~\ref{tab:NSE_resolution}.}
    \label{fig:ES}
\end{figure}

Fig.~\ref{fig:ES} presents the energy spectrum of data of initial condition from 1-D viscous Burgers' equation and 2-D Navier--Stokes equation with respect to different resolution settings. We can see in Fig.~\ref{fig:ES}(a) that the energy spectrum of the viscous Burgers' equation is similar for most of the wave numbers with different spatial resolutions. However, Fig.~\ref{fig:ES}(b) shows more noticeable differences in the energy spectrum of Navier--Stokes equations across the whole range of wave numbers with respect to different spatial resolutions. Unlike spatial discretization of VBE, where all the resolution settings provide a consistent result of the energy spectrum, the coarse resolution settings of NSE can cause over-estimations of energy in low wave numbers, which indicates that the numerical simulations do not well capture the true dynamical operator. Therefore, the trained models in the VBE example can approximate the true continuous operator and adapt well to the different resolution settings in Table.~\ref{tab:VBE_resolution}, with small test errors when making predictions in a higher spatial resolution. However, in the example of NSE, the trained models from coarse resolution settings in Table~\ref{tab:NSE_resolution} provide large test errors when making predictions in a higher spatial resolution, mainly due to the information loss in high wave numbers that prevent a good approximation of the true continuous operator.

\begin{figure}[H]
    \centering
    \includegraphics[width=0.9\textwidth]{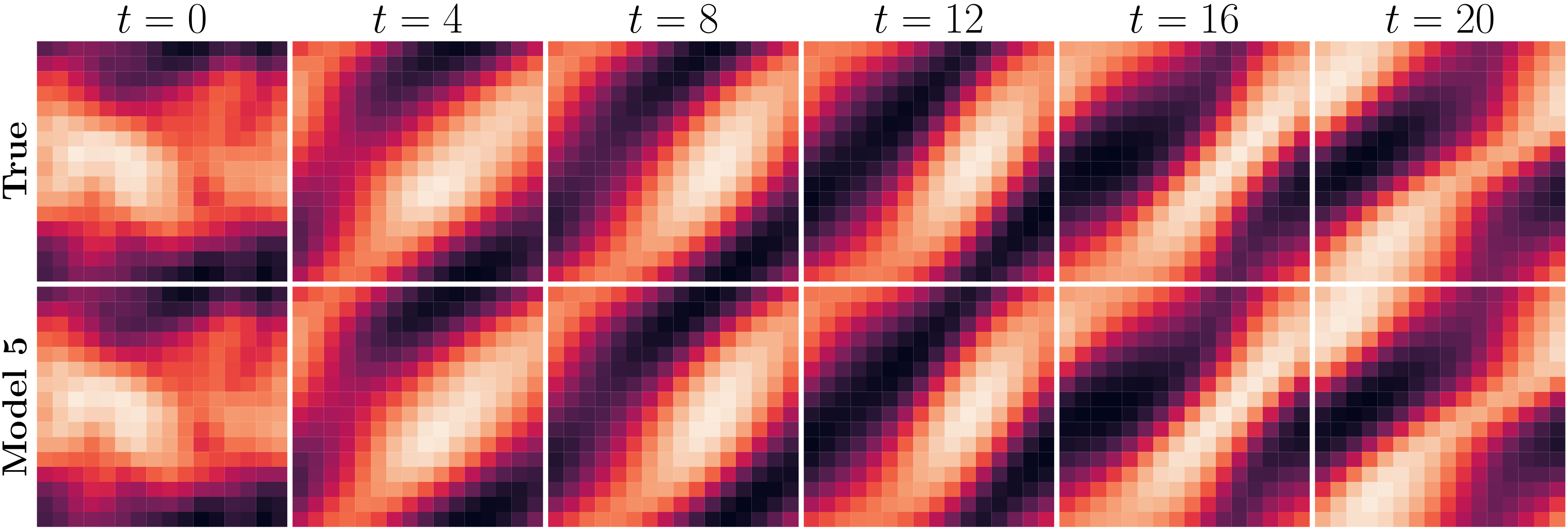}
    \caption{The spatial-temporal true simulation and model prediction. Upper row: true system with spatial resolution $16\times 16$. Lower row: predictions made by models trained with the data in a spatial resolution $16\times 16$ and tested on the data in the same resolution.}
    \label{fig:NSE_heatmap2}
\end{figure}

In Fig.~\ref{fig:NSE_heatmap2}, we present the spatial-temporal plots of the solutions with a spatial resolution $dx=1/16, dy=1/16$ for true system and the model trained by Resolution 5 in Table~\ref{tab:NSE_resolution}, with the initial condition sampled from test data. The upper row presents the true solution of $\omega$. The lower row corresponds to the prediction from the trained model. We can see that the trained model can capture the overall pattern of the true system in the spatial resolution $16 \times 16$. The error between upper and lower in Fig.~\ref{fig:NSE_heatmap2} corresponds to the Test Error (II) with Resolution 5 in Table.~\ref{tab:NSE_resolution}. It should be noted that the models trained by other resolutions in Table~\ref{tab:NSE_resolution} also have good prediction results in the same resolution as the corresponding training data, which are omitted here for simplicity.

\begin{figure}[H]
    \centering
    \includegraphics[width=0.9\textwidth]{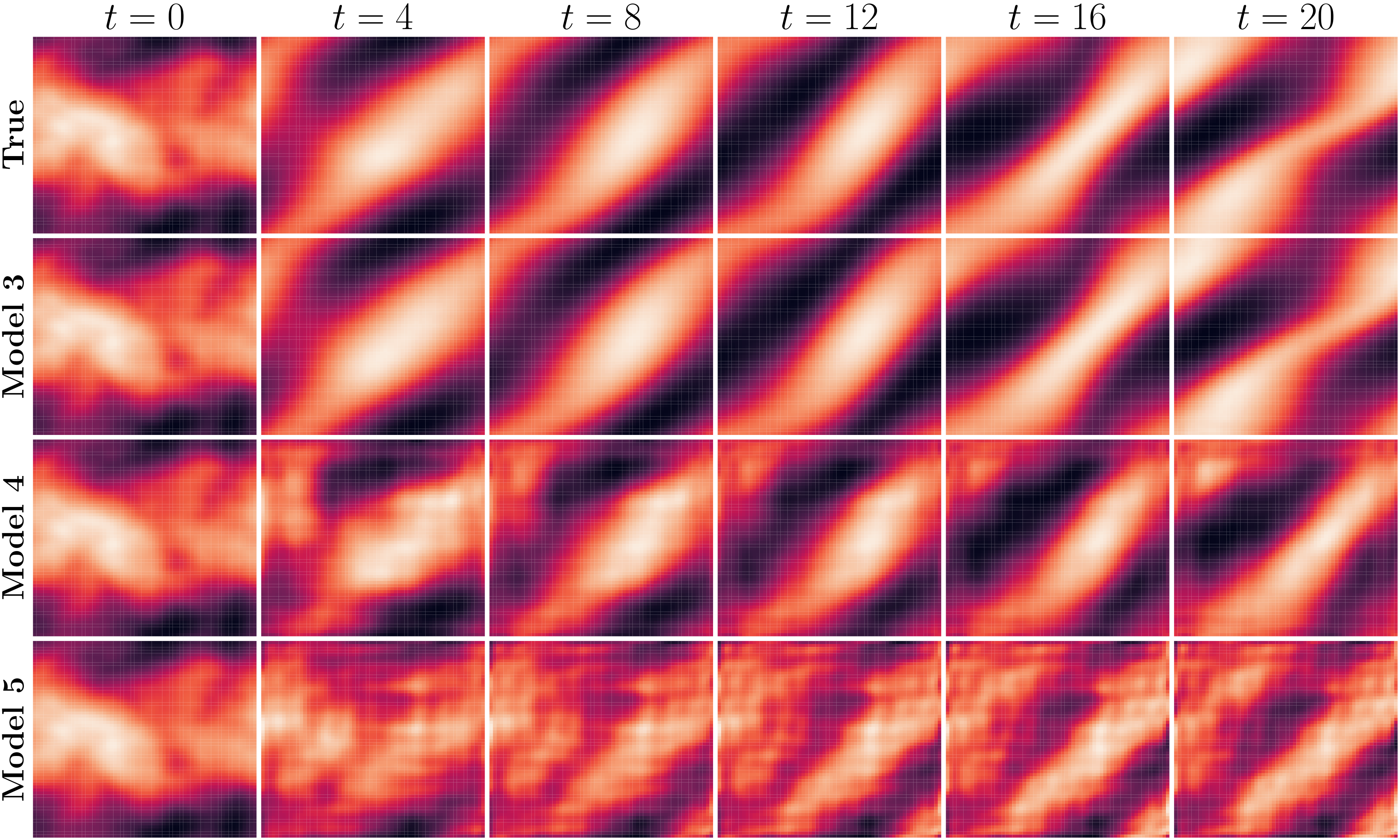}
    \caption{The flow of true simulation and model predictions of N-S equation from 0 to 20 time units. The predictions are made by models trained from different resolution settings in Table~\ref{tab:NSE_resolution}.}
    \label{fig:NSE_heatmap}
\end{figure}

We then present the spatial-temporal plots with the spatial resolution $dx=1/64, dy=1/64$ and the temporal resolution $dt=0.2$ in Fig.~\ref{fig:NSE_heatmap} for the true system and the prediction results of trained models. The initial condition is the same as the one used in Fig.~\ref{fig:NSE_heatmap2} but with a finer resolution $dx=1/64, dy=1/64$. The first row presents the true solution of $\omega$, and the other three rows correspond to the prediction results of the trained models with Resolutions 3, 4 and 5 in Table.~\ref{tab:VBE_resolution}. We can see that only Model 3 in Fig.~\ref{fig:NSE_heatmap} can capture the flow pattern of the true system, while Model 5 displays a noticeable mismatch with the true solution. Compared with Fig.~\ref{fig:NSE_heatmap2}, we can see that the results of Model 5 show a good performance for test data in spatial resolution $dx=1/16, dy=1/16$ (which is the same as train data), while the prediction results are unsatisfactory for the spatial resolution $dx=1/64, dy=1/64$. On the other hand, the good prediction results of Model 3 confirm that the trained model is temporal-invariant, i.e. capable of adapting to data with different temporal resolutions even with relatively sparse spatial data.

\begin{figure}[H]
    \centering
    \includegraphics[width=1\textwidth]{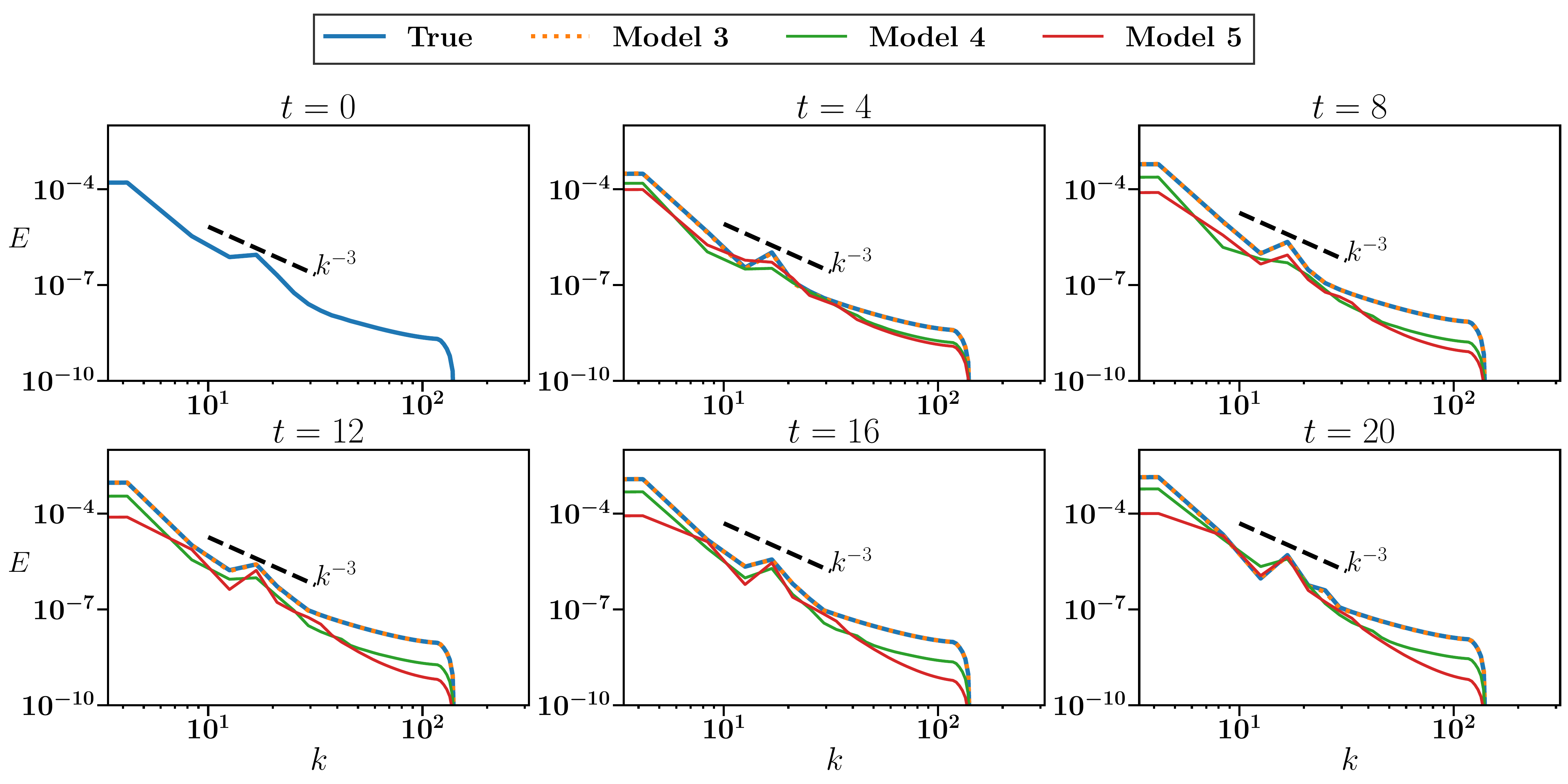}
    \caption{Energy spectrum of 2-D Navier--Stokes equation simulated with an initial condition from test data at various times.}.
    \label{fig:NSE_es}
\end{figure}

We further compare the energy spectrum of the solutions between the true system and the trained models in Fig.~\ref{fig:NSE_es}. A reference slope $k^{-3}$ is also included, which corresponds to the empirical decay rate of 2-D turbulence based on experimental data. The three trained models are tested with the same spatial-temporal resolution (i.e., $dx=1/64, dy=1/64, dt=0.2$).  We can see that only the energy spectrum of Model 3 has a good agreement with the true spectrum, while the results of Models 4 and 5 both demonstrate noticeable differences from the true one. It should be noted that the energy spectrum of the prediction results from Models 4 and 5 also do not agree well with the true results in Fig.~\ref{fig:ES}(b), which indicates that the trained dynamical operator only based on data in very coarse resolutions may not generalize well to the finer resolutions.

\subsection{Kuramoto–Sivashinsky Equation}
\label{ssec:KSE}

 Kuramoto–Sivashinsky (K-S) equation~\cite{kuramoto1976persistent, ashinsky1988nonlinear, hyman1986kuramoto, kevrekidis1990back} is a fourth-order nonlinear partial differential equation that was originally developed to model diffusive–thermal instabilities in a laminar flame front and features chaotic behavior and rich dynamics, e.g., dissipation and dispersion. The governing equation of the K-S equation is:  

\begin{equation}
\begin{aligned}
    \frac{\partial u}{\partial t} = -u\frac{\partial u}{\partial x} - \frac{\partial^2 u}{\partial x^2} - \frac{\partial^4 u}{\partial x^4},
\end{aligned}
\label{eq:KSE}
\end{equation}
where $x \in (0, L_x)$ with periodic boundary conditions, $t \in [0, L_t]$ and $u(x,0)$ is the given initial condition. We aim to learn a neural dynamical operator to approximate the right-hand-side of Eq.~\eqref{eq:KSE}, i.e., $\mathcal{G}: u \mapsto \frac{\partial u}{\partial t}$. In this example, we demonstrate that (i) the trained neural dynamical operator is spatial-temporal resolution-invariant and can provide good short-term prediction, (ii) the trained model based on short-term data can provide a stable long-term simulation with the chaotic behavior being retained qualitatively, and (iii) the model can achieve good performance for both short-term trajectory prediction and long-term statistics matching when trained with a hybrid optimization method. These three features are described in detail in the subsequent sections.

The simulation settings are $L_x=22, L_t=5000, dx=22/1024, dt=0.025$ and the initial condition is $u(x,0) = 0.1\times\cos(x/16)\times(1+2\sin(x/16))$. We simulate the true system with a single long trajectory, and the first $80\%$ of the trajectory (4000 time units) is used as train data and the remaining $20\%$ (1000 time units) is used as test data. For the K-S equation, 1000 time units is long enough to demonstrate the stability of long-term prediction of neural dynamical operator.

\subsubsection{Short-Term Prediction with Spatial-Temporal Resolution-Invariant}
We first train the model for short-term state prediction by solving the optimization problem in Eq.~\eqref{eq:opt}. Two short-term sub-trajectories of $20$ time units will be sampled from train time series data to serve as one data batch. The neural dynamical operator $\Tilde{\mathcal{G}}$ is constructed by a FNO model with $d_v=64$ and $k_{\textrm{max}}=24$. We train the model with $2\times 10^4$ epochs, and the optimizer is Adam with a learning rate $10^{-3}$ and cosine annealing schedule.

We train the neural dynamical operator based on short-term time series data with various resolutions in both space and time. The test results are summarized in Table \ref{tab:KSE_resolution}. The absolute test error is the mean squared error between true and predicted values for $20$ time units in test data. The long-term $D_{\textrm{KL}}$ is the Kullback–Leibler (KL) divergence (defined in Eq.~\eqref{eq:DKL}), which quantifies how a probability distribution differs from the other. Given PDF $p(x)$ from true data and $\Tilde{p}(x)$ from predicted data, the formula for the forward KL divergence of distributions $\Tilde{p}(x)$ from $p(x)$ is: 

\begin{equation}
\begin{aligned}
    D_{\textrm{KL}}(p || \Tilde{p}) = \int_{-\infty}^{\infty} p(x)\log(\frac{p(x)}{\Tilde{p}(x)})dx,
\end{aligned}
\label{eq:DKL}
\end{equation}
which is estimated from samples of both distributions based on k-Nearest-Neighbours probability density estimation \cite{wang2009divergence}. The forward KL divergence in Eq.~\eqref{eq:DKL} is mean-seeking, while the reverse KL divergence $D_{\textrm{KL}}(\Tilde{p} || p)$ is mode-seeking. In this work, both forward and reverse KL divergences are estimated based on 1000 time units in test data. Each point in Fig.~\ref{fig:KSE_DKL} stands for $D_{\textrm{KL}}$ from PDF of true data and PDF of predicted data.

We summarize the short-term test errors and long-term forward KL divergence for the system state $u$ in Table~\ref{tab:KSE_resolution}. The Test Error (I) and Long-Term $D_{\textrm{KL}}$ (I) are based on the test data of same resolution $dx=1/1024$ and $dt=0.05$, and the Test Error (II) and Long-Term $D_{\textrm{KL}}$ (II) is based on a resolution setting the same as the one of each train data. By comparing these test errors for trained models on different resolutions, it can be seen that the test error stays at the same order of magnitude when testing on a finer resolution in both space and time, confirming the resolution-invariance property of the trained models. In addition, all trained models lead to stable long-term simulations, which is mainly due to the high wavenumber filtering at each time step of evaluating the neural dynamical operator. More importantly, the stable long-term simulations of the trained models demonstrate small errors in KL divergence of the system state $u$, indicating a good quantitative agreement for the long-term prediction of the system state with the initial conditions from the test data. The visualization of $D_\textrm{KL}$ (I) is shown in the first plot of Fig.~\ref{fig:KSE_PDFACF}. It can be observed that the probability density functions (PDFs) obtained from models trained with the three different resolutions are very similar.

\begin{table}[H]
\begin{center}
\caption{The test errors of Kuramoto–Sivashinsky equation with various resolution settings for train data. The Test Error (I) and Long-Term $D_{\textrm{KL}}$ (I) are based on the test data of the resolution $dx=22/1024$ and $dt=0.25$, while the Test Error (II) and Long-Term $D_{\textrm{KL}}$ (II) are based on a resolution setting the same as each training data.}
\label{tab:KSE_resolution}
\begin{adjustbox}{max width=1.\textwidth, center}
\begin{tabular}{|c|cc|cc|c|cc|c|}
\hline
\multicolumn{1}{|c|}{\multirow{2}{*}{\diagbox[width=8em]{Resolution}{Error}}}  & \multicolumn{2}{c|}{Train Data}  & \multicolumn{2}{c|}{Test Error (I)} & Long-Term  &\multicolumn{2}{c|}{Test Error (II)} &Long-Term \\ \cline{2-5} \cline{7-8} 

\multicolumn{1}{|c|}{} & \multicolumn{1}{c|}{$dx$} & $dt$  & \multicolumn{1}{c|}{Absolute} & \multicolumn{1}{c|}{Relative} & \multicolumn{1}{c|}{$D_{\textrm{KL}}$ (I)} & \multicolumn{1}{c|}{Absolute} & \multicolumn{1}{c|}{Relative} & $D_{\textrm{KL}}$ (II) \\ \hline

Resolution1 & \multicolumn{1}{c|}{22/1024}  & 0.5 & \multicolumn{1}{c|} {4.6195e-02}  & 1.3609e-01  & 1.6776e-03 & \multicolumn{1}{c|}{4.5221e-02} & 1.3442e-01 & 1.6064e-03 \\ \hline

Resolution2 & \multicolumn{1}{c|}{22/512}  & 1 & \multicolumn{1}{c|}{6.3813e-02} & 1.4419e-01 & 5.1330e-03 & \multicolumn{1}{c|}{5.9550e-02} & 1.3860e-01 & 1.5060e-03 \\ \hline

Resolution3 & \multicolumn{1}{c|}{22/256}  & 2 & \multicolumn{1}{c|}{5.6369e-02} & 1.2588e-01 & 8.6506e-03 & \multicolumn{1}{c|}{4.9447e-02} & 1.1480e-01 & 1.7834e-03 \\ \hline
\end{tabular}
\end{adjustbox}
\end{center}
\end{table}

The Long-Term $D_{\textrm{KL}}$ (I) and (II) listed in the Table~\ref{tab:KSE_resolution} are the forward KL divergence. The corresponding reverse KL divergence for each resolution have also been calculated. The reverse Long-Term $D_{\textrm{KL}}$ (I) are 3.5586e-03, 6.2831e-03, 8.4653e-03 and reverse Long-Term $D_{\textrm{KL}}$ (II) are 3.3100e-03, 4.9920e-03, 1.6984e-03.

\begin{figure}[H]
    \centering
    \includegraphics[width=1\textwidth]{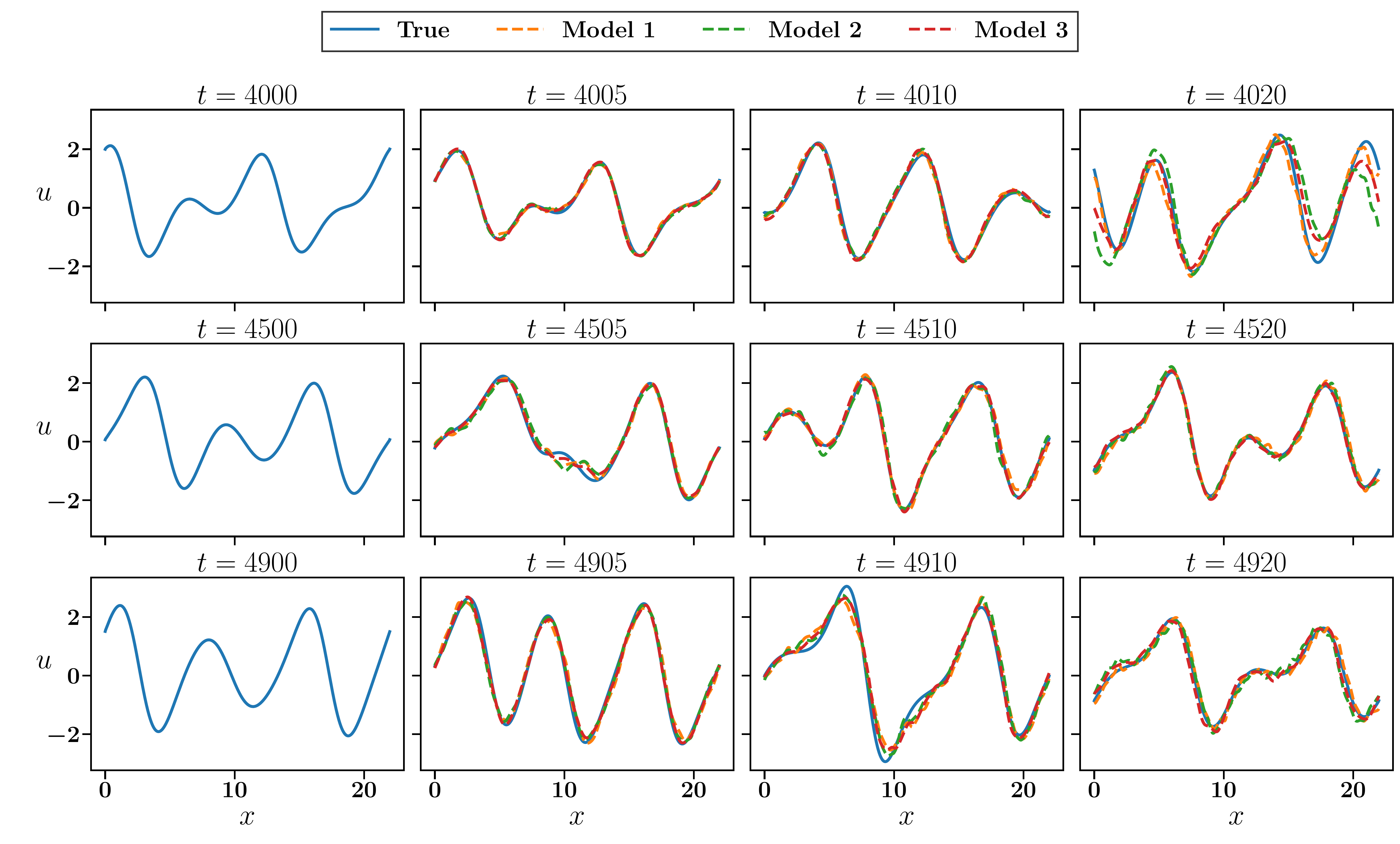}
    \caption{Solution profiles of the K-S equation for the true system and the model ones with different initial conditions from test data. The model is trained with each resolutions in Table.~\ref{tab:KSE_resolution} and tested on a finer resolution ($dx=22/1024, dt=0.25$).}
    \label{fig:KSE_profile}
\end{figure}

To facilitate a more detailed comparison between the short-term simulations of the true system and the modeled ones, we present the solution profiles $u$ in Fig.~\ref{fig:KSE_profile} for $20$ time units with three initial conditions (i.e., $t_0=4000, 4500, 4900$) sampled from the test data. Each row corresponds to a different test initial condition, and each model corresponds to a resolution setting with the same index in Table~\ref{tab:KSE_resolution}. The trained models are tested with a finer resolution $dx=22/1024$ and $dt=0.25$. We can see that the solution profiles of the trained model at various times all have a good agreement with the true solution profiles, with only some small deviation at a few regions. The results in Figs.~\ref{fig:KSE_pcolor} and~\ref{fig:KSE_profile} confirm that the trained neural dynamical operator has resolution-invariance in both space and time and also generalizes well to initial conditions from the test data.

\subsubsection{Stable Long-Term Simulation with Chaotic Behavior}
\begin{figure}[H]
    \centering
    \includegraphics[width=1\textwidth]{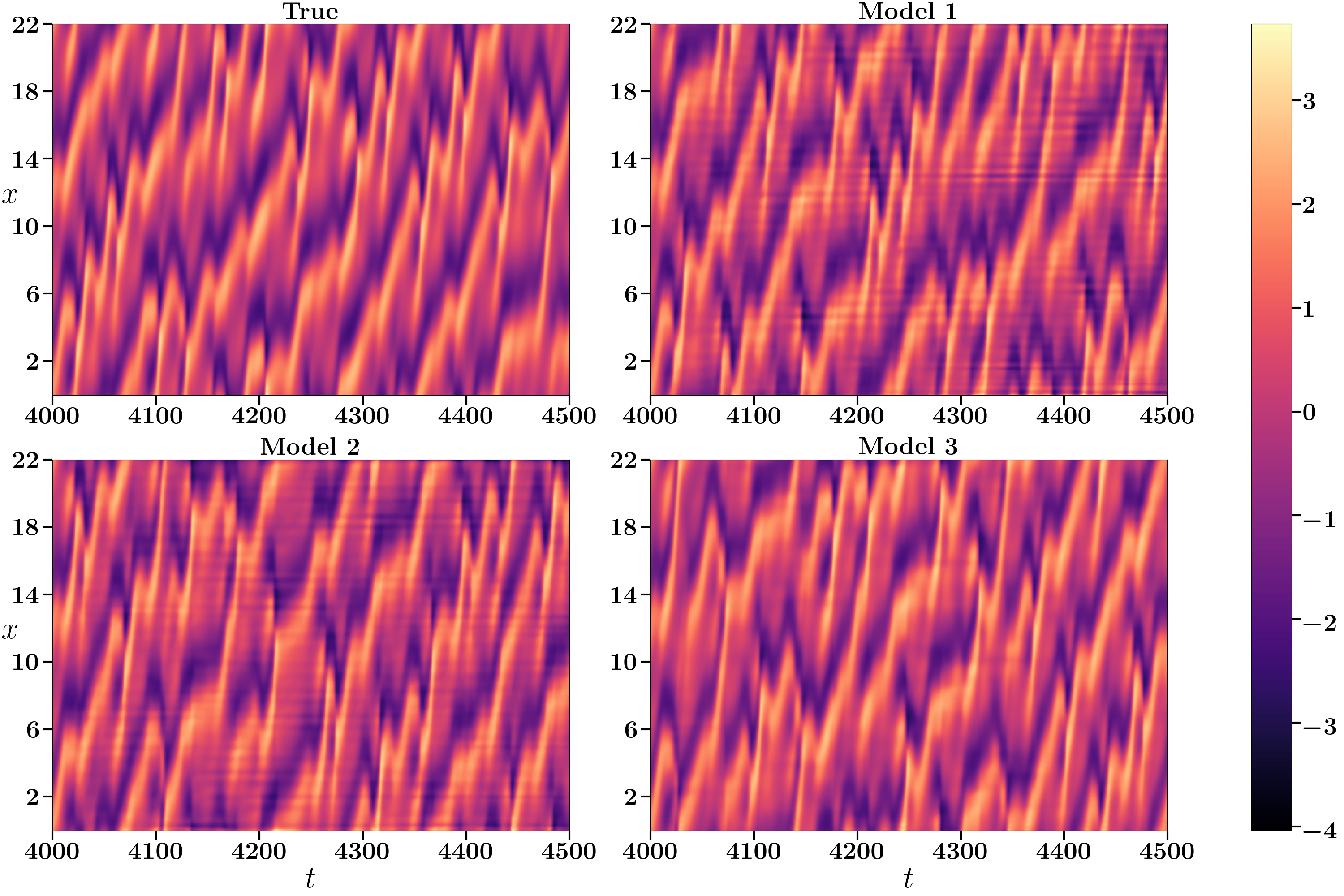}
    \caption{The 500s spatial-temporal solutions of Kuramoto–Sivashinsky equation. True: the solution simulated from the true system. Model: trained models from three different resolutions with the same test data resolution ($dx=22/1024, dt=0.25$). The index in the three trained models corresponds to the resolution settings in Table~\ref{tab:KSE_resolution}.}
    \label{fig:KSE_pcolor}
\end{figure}

In Fig.~\ref{fig:KSE_pcolor}, we present $500$ time units of spatial-temporal solutions from the true system and the modeled ones trained by the short-term time series data ($20$ time units), with the initial condition sampled from the test data. Considering that the K-S equation is a chaotic system, we would not expect a good quantitative agreement of long-term trajectories between the true system and the modeled one. It can be seen in Fig.~\ref{fig:KSE_pcolor} that the patterns of $500$ time units spatial-temporal solution plots demonstrate a good qualitative agreement with the true system, even though the models are trained with much shorter trajectories of the system state.

\begin{figure}[H]
    \centering
    \includegraphics[width=1\textwidth]{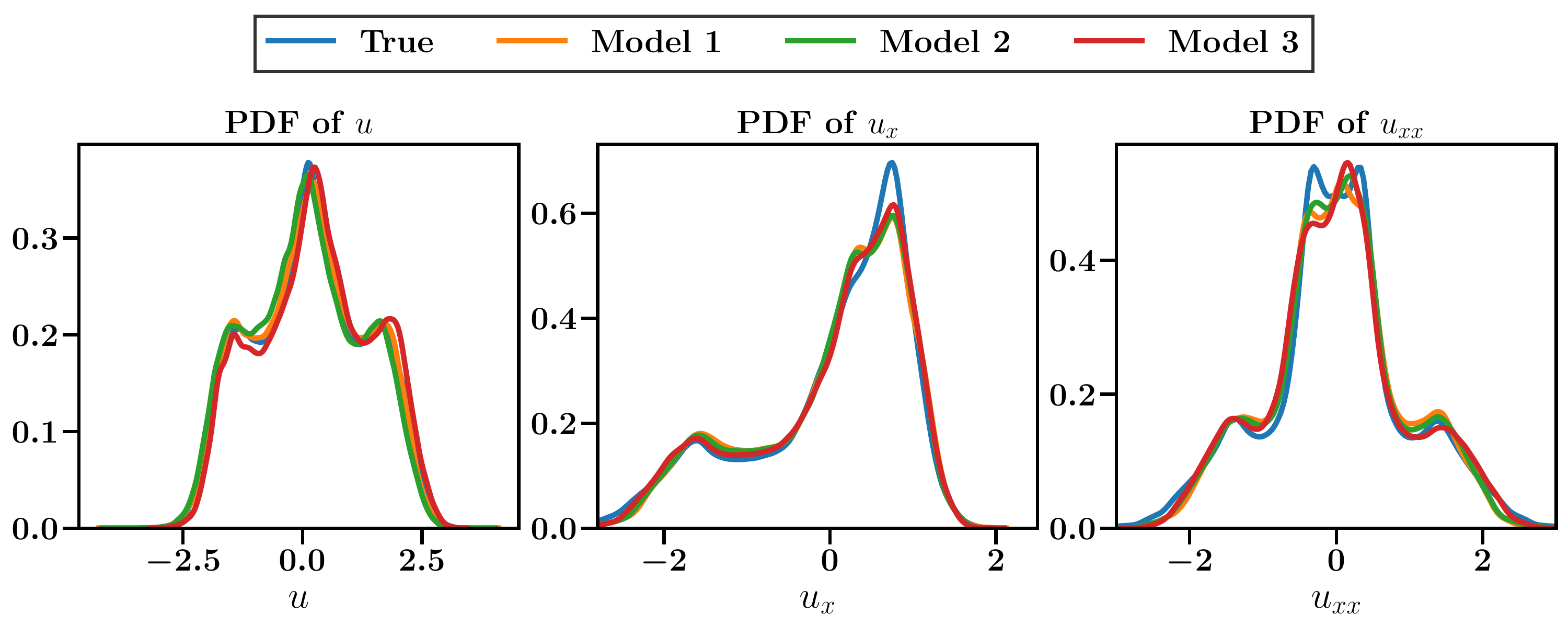}
    \includegraphics[width=0.85\textwidth]{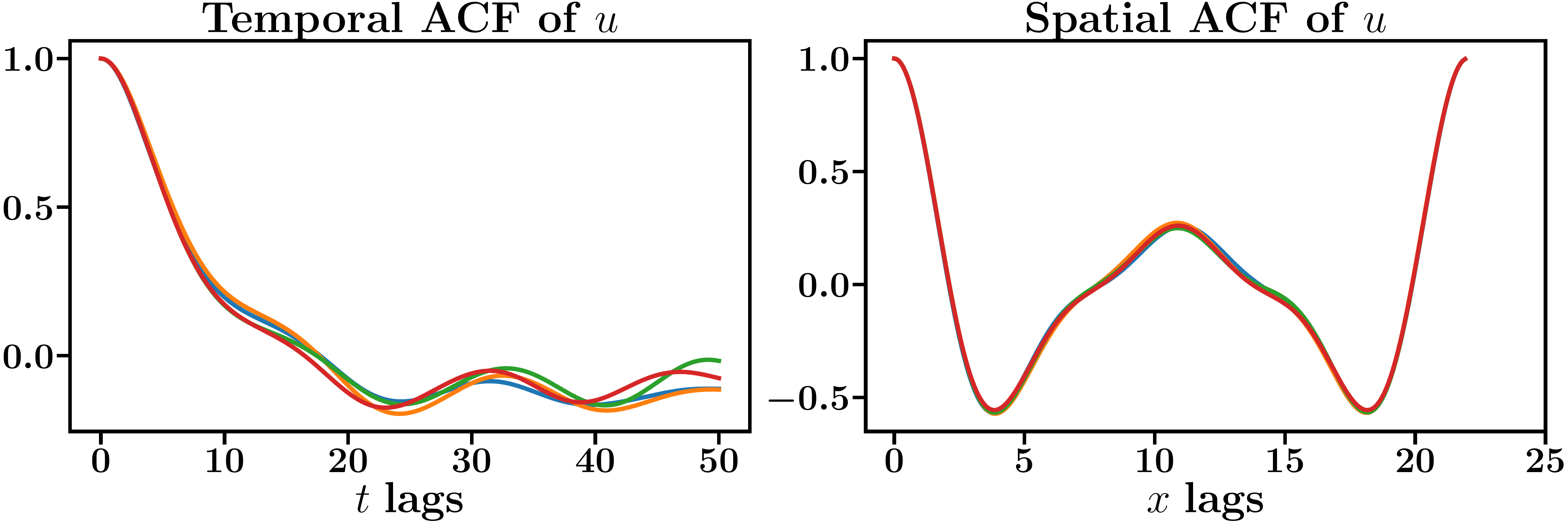}
    \caption{Probability density function and auto-correlation function of long-term (1000 time units) true simulation and model predictions for Kuramoto–Sivashinsky equation in test data. Upper: probability density function of state $u$, first spatial derivative $u_x$ and second spatial derivative $u_{xx}$. Below: temporal and spatial auto-correlation function of state $u$.}
    \label{fig:KSE_PDFACF}
\end{figure}

\begin{figure}[H]
    \centering
    \includegraphics[width=1\textwidth]{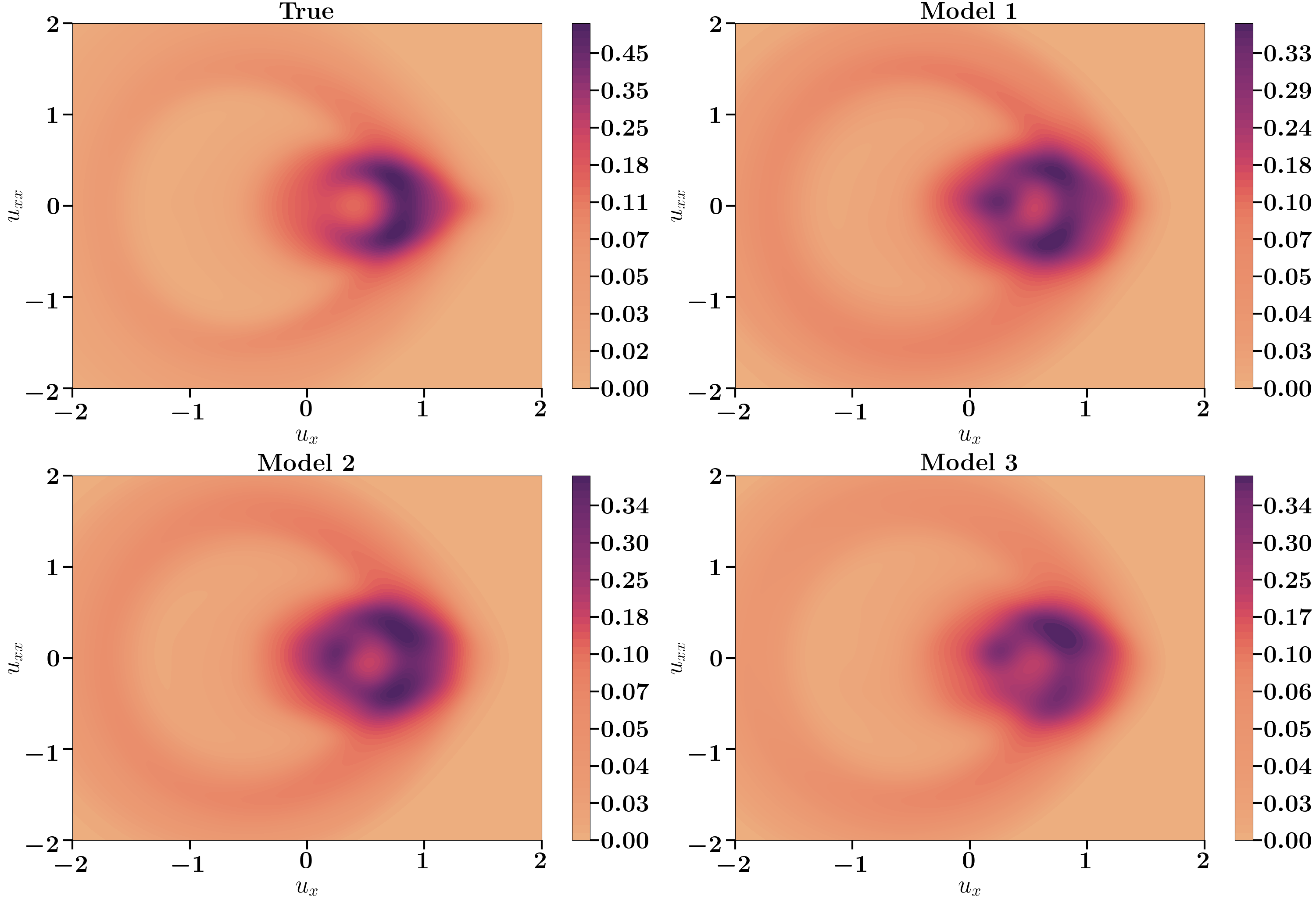}
    \caption{Joint probability density function of first spatial derivative and second spatial derivative $(u_x, u_{xx})$ for long-term (1000 time units) simulations from true and modeled systems in test data. The model systems are trained by resolution settings in Table.~\ref{tab:KSE_resolution} and tested on the resolution $dx=22/1024, dx=0.25$. }
    \label{fig:KSE_jointPDF}
\end{figure}

We further examine the statistical properties of long-term solutions from true and model systems in Fig.~\ref{fig:KSE_PDFACF}. Based on the true simulations and predictions for $1000$ time units, we compare the probability density function (PDF) of the system state $u$, first-order spatial derivative $u_x$, second-order spatial derivative $u_{xx}$, and spatial and temporal auto-correlation function (ACF) of $u$. We can find that the PDF of state $u$ and its first spatial derivative from long-term prediction can match the true simulation well, while the PDF of the second spatial derivative $u_{xx}$ from the modeled systems shows a less satisfactory agreement with the true one. Also, the temporal and spatial ACF of state $u$ show a similar pattern between predicted and true values, indicating a stable long-term prediction by trained models with similar statistical properties.  In Fig.~\ref{fig:KSE_jointPDF}, we also show the joint probability density function of $(u_{x}, u_{xx})$ for the long-term (1000 time units) simulation from true system and modeled systems with the same resolution $dx=22/1024,dt=0.25$. Compared with the joint PDF from true simulation, even though the joint PDFs from model predictions have a relatively lower maximum density and spread more out, their overall patterns are still qualitatively similar to the true system.

\begin{figure}[H]
    \centering
    \includegraphics[width=\textwidth]{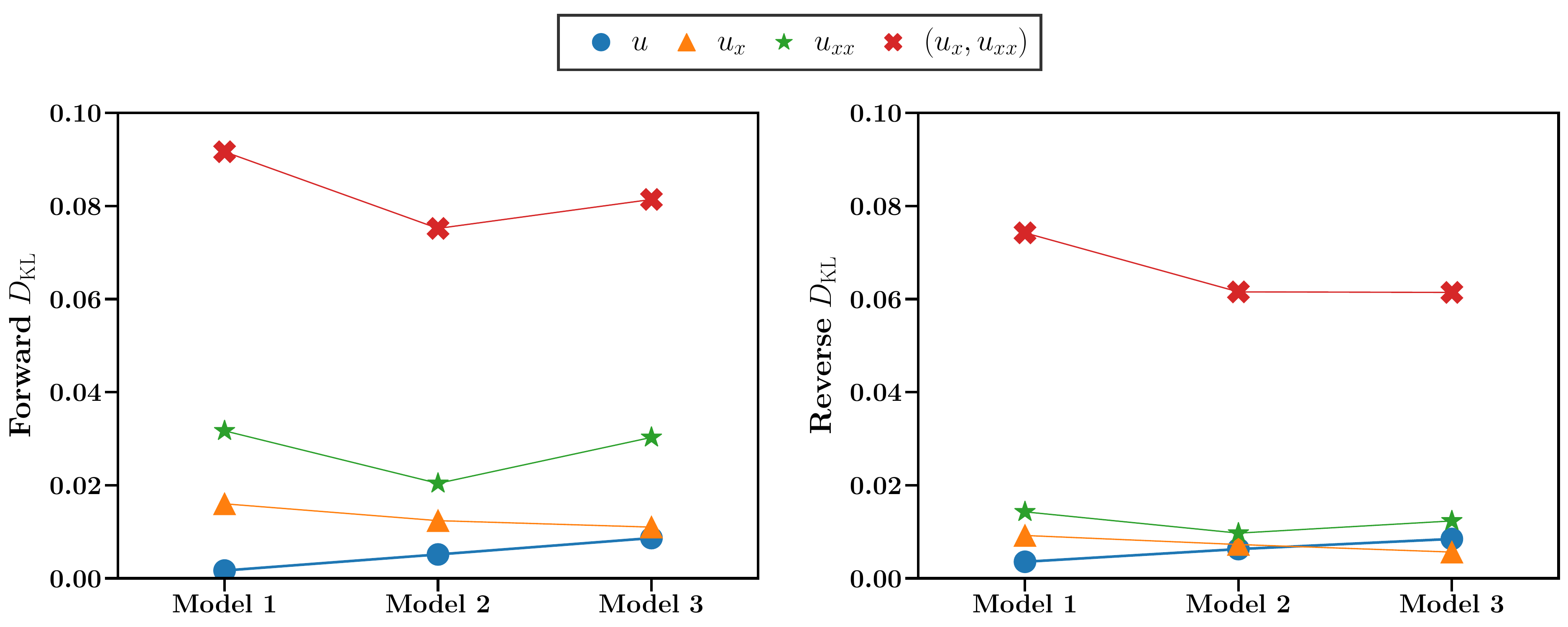}
    \caption{Summary of KL Divergence between PDFs from 1000 time units simulation from true system and modeled system. Those PDFs and joint PDF includes $u$, $u_x$, $u_{xx}$, and $(u_x, u_{xx})$.}
    \label{fig:KSE_DKL}
\end{figure}

Besides the qualitative visualization of those probability density functions, the KL divergence from PDFs of true data to PDFs of model predictions are calculated and summarized in Fig.~\ref{fig:KSE_DKL}. The left panel displays the forward KL divergence, while the right panel shows the reverse KL divergence. The KL divergence results include $u$, $u_x$, $u_{xx}$ and ($u_x$, $u_{xx}$) and show small values for all the trained models with different training resolutions, demonstrating the resolution-invariance property of trained neural dynamical operator even in long-term predictions.

\subsubsection{Hybrid Optimization for Short-Term State Prediction and Long-Term Statistics Matching}

The previous two sections show that the neural dynamical operator, trained with abundant short-term data, can make accurate short-term state predictions and consistent long-term statistics with the feature of resolution-invariant if sufficient time series data (4000 time units) is available. To demonstrate the merits of the hybrid optimization, we consider a practical application with data scarcity: only some short-term batches and long-term statistics from the 4000 time units training data in resolution 3 ($dx=22/256, dt=2$) are available. More specifically, the data available for use include 40 short-term batches, each with a length of 2 time units, and 20 sets of long-term statistics (variance) from 200 time units with known initial conditions distributed evenly through training data. Trained with only the short-term batches, the neural dynamical operator can still produce relatively accurate short-term state prediction and stable long-term simulation. However, the long-term variance is much larger than the truth. With the data of long-term variance, hybrid optimization can effectively calibrate the pre-trained neural dynamical operator to enhance the consistency with the long-term statistics of the true system. The forward map in Eq.~\eqref{eq:EKI_G} of EKI is constructed by the composition of three components: (i) generating long-term simulations of 200 time units with neural dynamical operator, (ii) calculating the second-order spatial derivative $u_{xx}$ from the simulated system state $u$, and (iii) calculating the variance of $u_{xx}$. 

The main reason for starting from the model trained with short-term data is for efficiency consideration. More specifically, the long-term statistics engaged in this example is a scalar quantity, while the short-term loss aims to match a trajectory. Therefore, the amount of data is larger in the short-term loss than in the long-term one. Considering that the two losses are iteratively optimized in the hybrid algorithm, engaging the long-term loss at the beginning of the whole optimization process is not efficient and could have a negative impact on the convergence of the iterative optimization algorithm, as the information in the long-term statistics is far from enough to help guide the randomly initialized model to gradually mimic the true system. This example represents a large class of real-world applications, whose available long-term statistics are not as informative as the short-term trajectory data but still significant. In this work, we engage the long-term statistics via EKI as a modification of the pre-trained model with short-term loss only.

Starting with the pre-trained model, the hybrid optimization updates the neural dynamical operator by alternating the short-term trajectory matching via gradient-based optimization (i.e., Adam optimizer in this work) and the long-term statistics matching via derivative-free optimization (i.e., EKI in this work). More details about the hybrid optimization algorithm have been summarized in Algorithm~\ref{alg:HybridOptimization}. The number of training epochs $N_{\textrm{ep}}$ of the gradient-based optimization is chosen as 3000 and the number of training epochs of EKI is chosen as 10, which means that all the trainable parameters in the neural dynamical operator will be updated by one EKI training epoch after every 300 epochs of the gradient-based optimization. The learning rate of the Adam optimizer is set as $10^{-4}$, considering that we are tuning a well-trained model at the beginning. In each EKI epoch, $N_{\textrm{it}}=20$ iterations will be performed with an ensemble size $J=100$. 

\begin{figure}[H]
    \centering
        \includegraphics[width=1\textwidth]{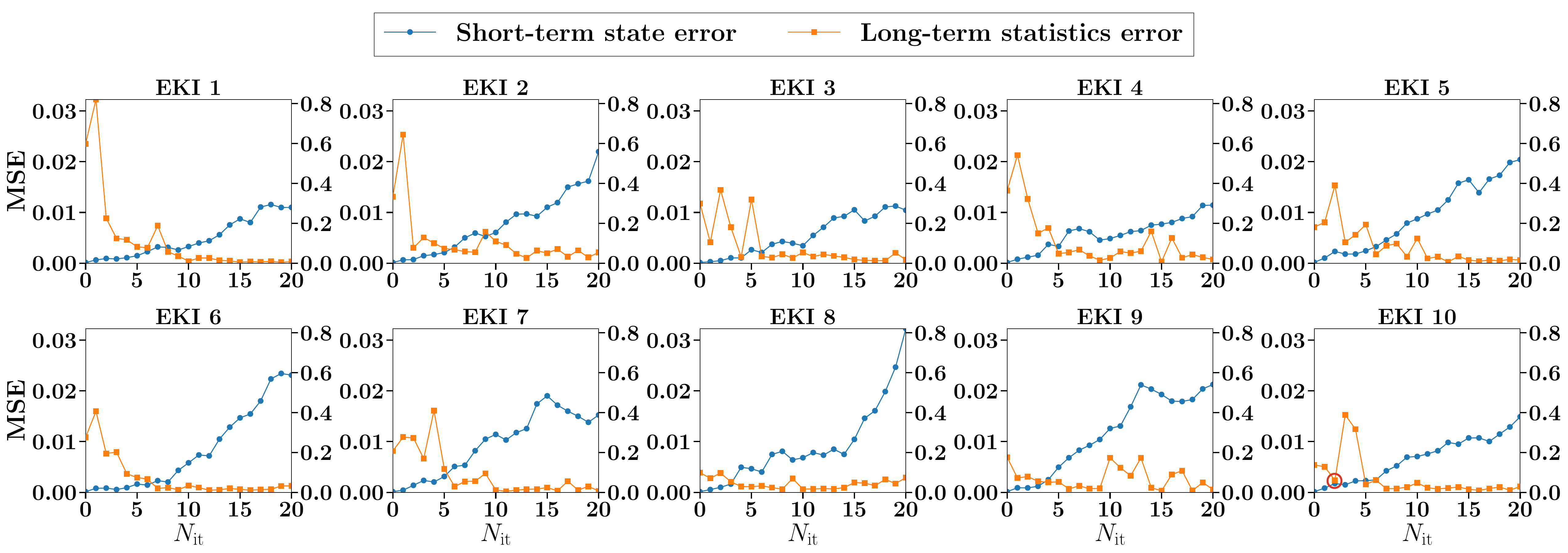}
            \caption{Long-term and short-term error history of the EKI epochs in the hybrid optimization. In each EKI epoch, the parameters will be updated 20 iterations based on Eq.~\eqref{eq:EKI_update}. The $0$-th iteration is the error of the model updated by the previous gradient-based optimization epochs. In each sub-figure, the right axis is the short-term state MSE, and left axis is the long-term statistics MSE.}
    \label{fig:KSE_EKI_Error}
\end{figure}

We present the error history during the EKI updating in Fig.~\ref{fig:KSE_EKI_Error}. The short-term error is the mean squared error for short-term ($2$ time units) system state trajectory, while the long-term error is the mean squared error of variances from long-term simulations with 200 time units. We can see that there is a trade-off between short-term state error and long-term statistics error in each EKI epoch, mainly because each EKI epoch only focuses on long-term statistics matching in the proposed hybrid optimization method. Although the short-term error tends to increase within each EKI epoch, the subsequent epochs of gradient-based optimization with short-term trajectory matching would keep tuning the neural dynamical operator such that a smaller short-term error is achieved. With the EKI loss training history, we selected the parameters trained after two iterations in $10^\mathrm{th}$ EKI epoch, highlighted by the red circle in the third column of Fig.~\ref{fig:KSE_EKI_Error}. It is worth noting that the proposed method sequentially optimizes the short-term loss $L_s$ and the long-term loss $L_l$ individually and then chooses a robust result, i.e., relatively small values of $L_s$ and $L_l$ compared to their range of values, respectively. In general, the approach is related to solving a two-objective optimization problem based on the Pareto front, which is estimated via iteratively solving the optimization problem that only involves $L_s$ or $L_l$.

\begin{figure}[H]
    \centering
    \includegraphics[width=1\textwidth]{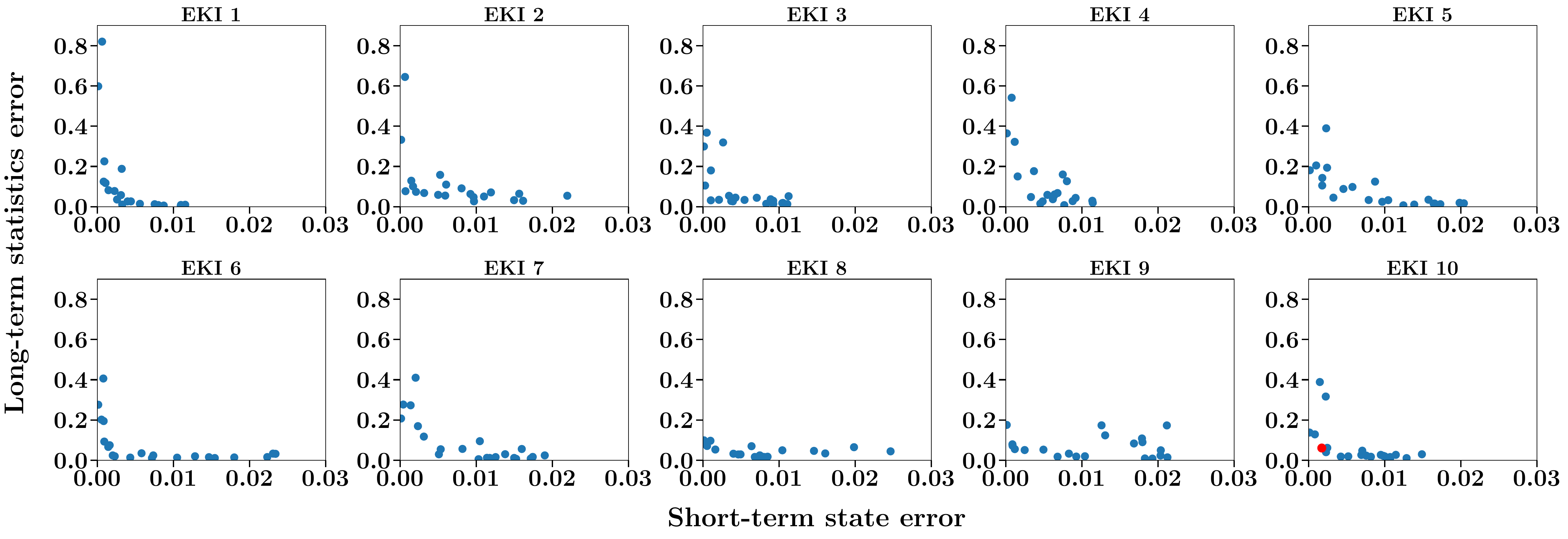}
    \caption{Long-term statistics error versus short-term state error during the EKI updating in the hybrid optimization. Each sub-figure corresponds to the counterparts in the Fig.~\ref{fig:KSE_EKI_Error}. }
    \label{fig:KSE_EKI_Error2}
\end{figure}

The relationship between long-term statistics error and short-term state error during the EKI training is presented in Fig.~\ref{fig:KSE_EKI_Error2}. Each sub-figure corresponds to its counterpart in Fig.~\ref{fig:KSE_EKI_Error}. In each EKI iteration, we observe a decreasing trend of long-term statistics error while an increase in short-term error. This again manifests the trade-off between long-term and short-term errors during the EKI updating process. In this section, we select the trained model that provides a balanced performance of short-term trajectory and long-term statistics, which corresponds to the red point of the $10^\mathrm{th}$ EKI epoch at the bottom-left region (i.e., relatively small short-term and long-term errors) in Fig.~\ref{fig:KSE_EKI_Error2}. The selected model is also highlighted by the red circle in Fig.~\ref{fig:KSE_EKI_Error}. Two other trained models with relatively small short-term and long-term errors are presented in \ref{sec:additional_results} to demonstrate the robustness of the proposed hybrid optimization approach.

\begin{figure}[H]
    \centering
    \includegraphics[width=1\textwidth]{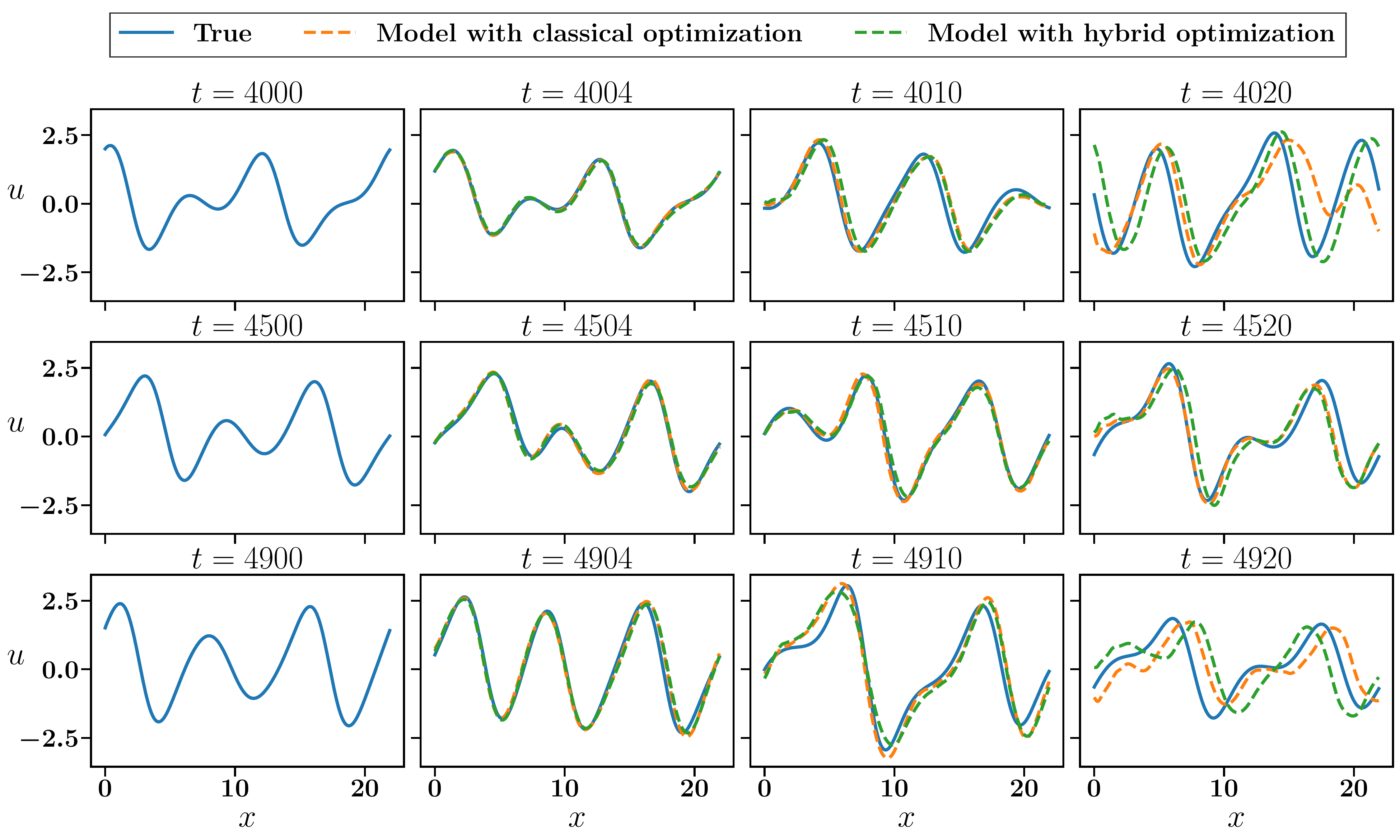}
    \caption{Solution profiles of the K-S equation for the true system, model trained with classical optimization, and model trained via hybrid optimization with different initial conditions in test data. The model with classical optimization are trained with 40 short-term batches with length of 2 time units from training data with resolution $dx=22/256, dt=2$. The model with hybrid optimization is further calibrated using the 20 sets of long-term statistics data. Both models are tested on the same resolution.}
    \label{fig:KSE_profile_hybrid}
\end{figure}

The short-term solution profiles from the simulation of the true system and predictions of models with classical and hybrid optimization schemes are presented in Fig.~\ref{fig:KSE_profile_hybrid}. The simulation of the true system has the resolution $dx=22/256, dt=2$, and both models are trained with the the 40 short-term batches with 2-time-unit length from the simulation. The three initial conditions are from test data the same as Fig.~\ref{fig:KSE_profile}. From the comparison of solution profiles starting with those initial conditions in Fig.~\ref{fig:KSE_profile_hybrid}, we can find that short-term predictions from both models are similar to the true solution profiles when the predictive horizon less than 20 time units. The model trained solely with classical optimization performs slightly better than the model with hybrid optimization. More specifically, the absolute and relative short-term state prediction errors for 20-time-units of the model with classical optimization are 0.2765 and 0.2865, while the errors of the model with hybrid optimization are 0.3601 and 0.3565, respectively.

\begin{figure}[H]
    \centering
    \includegraphics[width=0.5\textwidth]{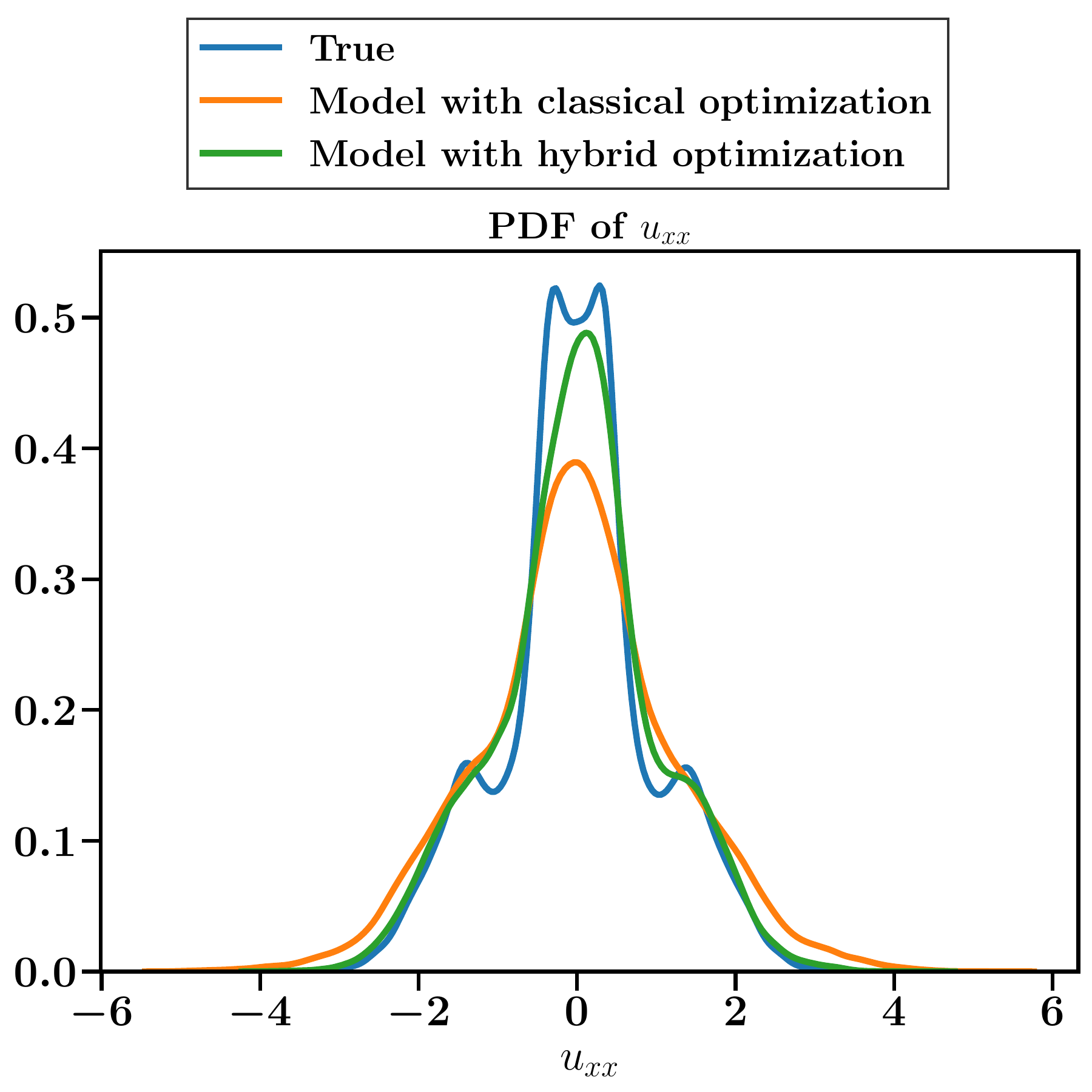}
    \caption{Probability density function of second spatial derivative $u_{xx}$ from long-term (1000 time units) simulation of true system, five sets of 200 time units simulation from model trained with classical optimization and model trained with hybrid optimization with initial conditions distributed evenly from test data.}
    \label{fig:KSE_uxx_hybrid}
\end{figure}

With a comparable performance of short-term prediction to the model with classical optimization, hybrid optimization can lead to better long-term statistics as presented in Fig.~\ref{fig:KSE_uxx_hybrid}. More specifically, the mean squared error of the long-term statistics (i.e., the variance of $u_{xx}$) predicted by the model with classical optimization is 0.5340, while the one with hybrid optimization is 0.0187. This indicates that the model with hybrid optimization has a much better agreement with the true value of the variance than the one with solely classical optimization. We present the probability density distribution of the second spatial derivative $u_{xx}$ for 1000 time units test data from the true system and the results of the two trained models in Fig.~\ref{fig:KSE_uxx_hybrid}. It should be noted that the PDFs from models are based on five sets of 200-time-unit simulations with initial conditions distributed evenly through the test data. The peak part of the PDF from the model with hybrid optimization is higher than the model with classical optimization, and its tail part is also more in line with the true PDF. This improvement contributed by the hybrid optimization approach indicates that the trained model is more capable of quantifying the dispersion and uncertainty of the true system, e.g., extreme events that are of interest in science (e.g., extreme weather/climate) and engineering (e.g., responses of materials or energy systems with extreme loads) applications. As shown in Fig.~\ref{fig:KSE_EKI_Error}, the highlighted point in the $10^\mathrm{th}$ is not the only choice of the trained model. To demonstrate the robustness of the proposed hybrid optimization approach, trained models correspond to some other points presented in~\ref{sec:additional_results}.

\section{Conclusion}
A recent trend of data-driven modeling is to formulate the problem in its continuous form, which facilitates a more flexible use of data in general. The merits of existing spatially continuous models (e.g., neural operator) and temporally continuous models (e.g., neural ODE) have been demonstrated in many science and engineering applications. In this work, we present a data-driven modeling framework that learns a continuous spatial-temporal model based on the techniques of neural operator and neural ODE. More specifically, we focus on the learning of the dynamical operator and demonstrate that the learned model is resolution-invariance in both space and time. We also show that the learned model can provide stable long-term simulations, even if the training data only contains short-term time series of true system states. In addition, we propose a hybrid optimization scheme that leverages both gradient-based and derivative-free methods and efficiently combines the use of short-term time series and long-term statistics in training the model. The proposed framework is studied based on three classical examples governed by partial differential equations, including the viscous Burgers' equation, the Navier--Stokes equation, and the Kuramoto--Sivashinsky equation. The results show that: (i) the trained model has resolution-invariance with respect to both spatial and temporal discretizations, and (ii) the hybrid optimization scheme ensures a good performance of the trained model in both matching short-term trajectories and capturing long-term system behaviors. Although this work mainly focuses on demonstrating the use of FNO to construct the neural dynamical operator and the iterative optimization of short-term and long-term losses, it is worth noting that other operator learning techniques (e.g., DeepONet) and optimization methods (e.g., $\epsilon-$constrained method for multiobjective optimization) can also be employed by the proposed framework.

\section*{Acknowledgments}
J.W. and C.C. are supported by the University of Wisconsin-Madison, Office of the Vice Chancellor for Research and Graduate Education with funding from the Wisconsin Alumni Research Foundation.

\section*{Data Availability}
The data that support the findings of this study are available from the corresponding author upon reasonable request. The codes and examples that support the findings of this study are available in the link: \url{https://github.com/AIMS-Madison/Neural-Dynamical-Operator}.


\clearpage
\appendix
\section{Hybrid Optimization for Neural Dynamical Operator}
\label{sec:algorithm}
The detailed algorithm of the hybrid optimization is presented in Algorithm~\ref{alg:HybridOptimization}. In this work, we apply this algorithm to the example of Kuramoto–Sivashinsky equation in Section~\ref{ssec:KSE}, for an efficient and robust training of neural dynamical operator with both short-term and long-term data.

\begin{algorithm}[H]
\caption{Training neural dynamical operator with the hybrid optimization scheme}
\label{alg:HybridOptimization}
\begin{algorithmic}[1]
\State \textbf{Input}:
\State \quad \quad $\{\boldsymbol{u}(t_n)\}_{n=0}^{N} \gets $ Training Data
\State \quad \quad $\{t_n\}_{n=0}^{N} \gets$  Time Stamp
\State \quad \quad $\tilde{\mathcal{G}}(\boldsymbol{\theta}) \gets $ Neural Operator
\State $(N_s, N_l) \gets (20, 1000)$
\State $(N_{\textrm{ep}}, k, \lambda, N_{\textrm{it}}, J)\gets (3000, 300, 1e^{-3}, 20, 100)$
\State $\boldsymbol{\Theta} \gets$  Empty List
\For { $i=1,2,..., N_{\textrm{ep}}$ } \Comment{\textbf{SGD Training}}
    \State $N_0 \sim \mathbb{U}(0, N - N_s)$
    \State $\{ \tilde{\boldsymbol{u}}(t_n) \}_{n=N_0}^{N_0 + N_s} = \text{ODESolver}(\tilde{\mathcal{G}}(\boldsymbol{\theta}), \boldsymbol{u}(t_{N_0}), \{t_n\}_{n=N_0}^{N_0 + N_s})$
    \State $L_s = \Sigma_{n=N_0}^{N_0 + N_s} || \boldsymbol{u}(t_n) - \tilde{\boldsymbol{u}}(t_n) ||^2$
    \State $\boldsymbol{\theta} \gets \boldsymbol{\theta} - \lambda \cdot \nabla_{\boldsymbol{\theta}} L_s$ \Comment{Parameters updated by SGD}
    \If { $i \bmod k =0 $ } \Comment{\textbf{EKI Training}}
        \State $N_0 \sim \mathbb{U}(0, N - N_l)$
        \State $\boldsymbol{y} = \beta( {\{ \tilde{\boldsymbol{u}}(t_n) \}_{n=N_0}^{N_0 + N_l}} )$ \Comment{Long-term Statistics}
        \State $\boldsymbol{\Sigma}_{\boldsymbol{\eta}} = 0.01\boldsymbol{\mathrm{I}}_{d_{\boldsymbol{y}}}$  
        \State $\{ \boldsymbol{\theta}^{(j)} \}_{j=1}^J = \boldsymbol{\theta} + \{\boldsymbol{\epsilon}^{(j)}\}_{j=1}^J, \quad \boldsymbol{\epsilon}^{(j)} \sim \mathcal{N}(0, 0.1^2\boldsymbol{\mathrm{I}}_{d_{\boldsymbol{\theta}}}) $
        \For {$m=1,2,... N_{\textrm{it}}$}
            \State $\{[\{ \tilde{\boldsymbol{u}}(t_n) \}_{n=0}^{N_l}]^{(j)}\}_{j=1}^J = \{\text{ODESolver}(\tilde{\mathcal{G}}( \boldsymbol{\theta}^{(j)} ), \boldsymbol{u}(t_{N_0}), \{t_n\}_{n=N_0}^{N_0 + N_l})\}_{j=1}^J$
            \State $\{\boldsymbol{g}^{(j)}\}_{j=1}^J = \{ \beta([\{ \tilde{\boldsymbol{u}}(t_n) \}_{n=0}^{N_l}]^{(j)})\}_{j=1}^J$  
            \State $\{\boldsymbol{y}^{(j)}\}_{j=1}^J = \boldsymbol{y} + \boldsymbol{\eta}^{(j)}, \quad \boldsymbol{\eta}^{(j)} \sim \mathcal{N}(0, \boldsymbol{\Sigma}_\eta)$
            \State $\{ \boldsymbol{\theta}^{(j)} \}_{j=1}^J \gets \{\boldsymbol{\theta}^{(j)} + \boldsymbol{\Sigma}^{\boldsymbol{\theta} \boldsymbol{g}}(\boldsymbol{\Sigma}^{\boldsymbol{gg}} + \boldsymbol{\Sigma}_{\boldsymbol{\eta}})^{-1}(\boldsymbol{y}^{(j)} - \boldsymbol{g}^{(j)}) \}_{j=1}^J$ \Comment{EKI updating formula}
        \EndFor
    \EndIf
    \State $\boldsymbol{\theta} = \frac{1}{ J}\Sigma_{j=1}^{J} \boldsymbol{\theta}^{(j)}$   \Comment{Parameters updated by EKI}
    \State Append $\boldsymbol{\theta}$ to $\boldsymbol{\Theta}$

\EndFor
\State \text{Select robust $\boldsymbol{\theta}^{\star}$ from $\boldsymbol{\Theta}$ based on the error history during EKI updating.}
\end{algorithmic}
\end{algorithm}

\section{Additional Results of Hybrid Optimization Approach}
\label{sec:additional_results}

\begin{figure}[H]
    \centering
    \includegraphics[width=\textwidth]{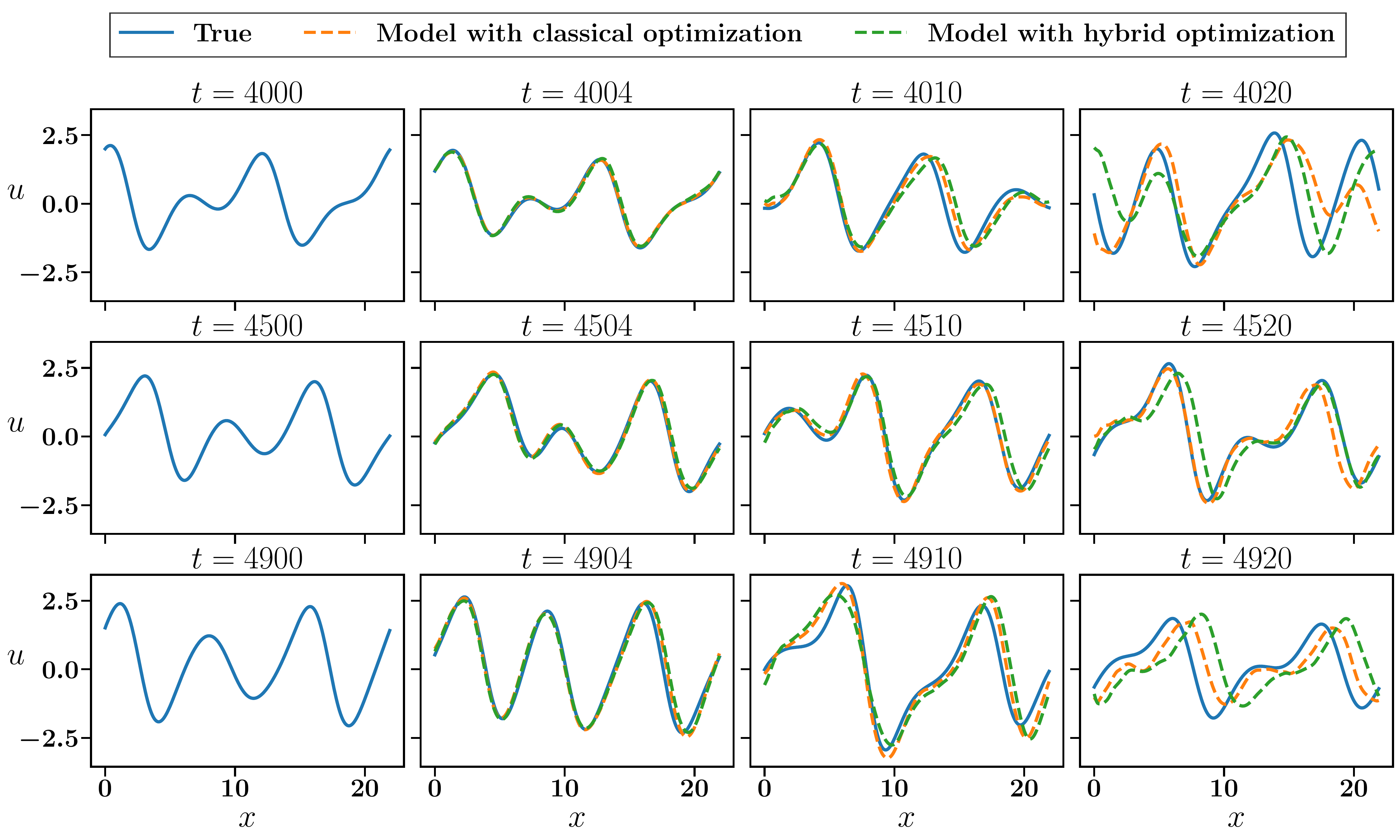}
    \caption{Solution profiles of the K-S equation for the true system, model trained with classical optimization, and model trained via hybrid optimization with different initial conditions in test data. The model with classical optimization are trained with 40 short-term batches with length of 2 time units from training data with resolution $dx=22/256, dt=2$. The model with hybrid optimization is further calibrated using the 20 sets of long-term statistics data. Both models are tested on the same resolution. The trained model via hybrid optimization corresponds to the point after one iteration in the $9^\mathrm{th}$ EKI in Fig.~\ref{fig:KSE_EKI_Error}.}
    \label{fig:KSE_profile_hybrid_1}
\end{figure}

\begin{figure}[H]
    \centering
    \includegraphics[width=0.5\textwidth]{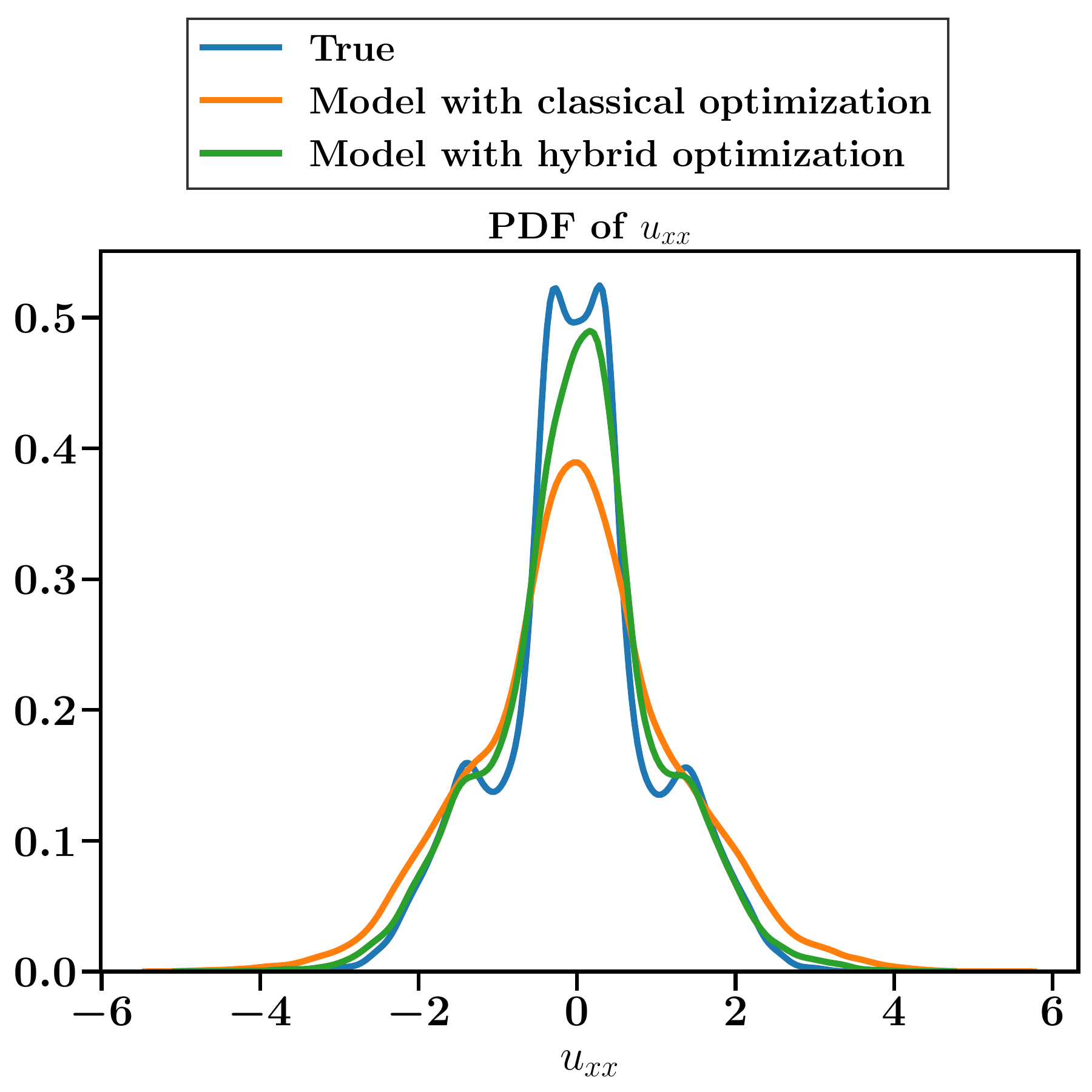}
    \caption{Probability density function of second spatial derivative $u_{xx}$ from long-term (1000 time units) simulation of true system, five sets of 200 time units simulation from model trained with classical optimization and model trained with hybrid optimization with initial conditions distributed evenly from test data. The trained model via hybrid optimization corresponds to the point after one iteration of the $9^\mathrm{th}$ EKI epoch in Fig.~\ref{fig:KSE_EKI_Error}.}
    \label{fig:KSE_uxx_hybrid_1}
\end{figure}

\begin{figure}[H]
    \centering
    \includegraphics[width=\textwidth]{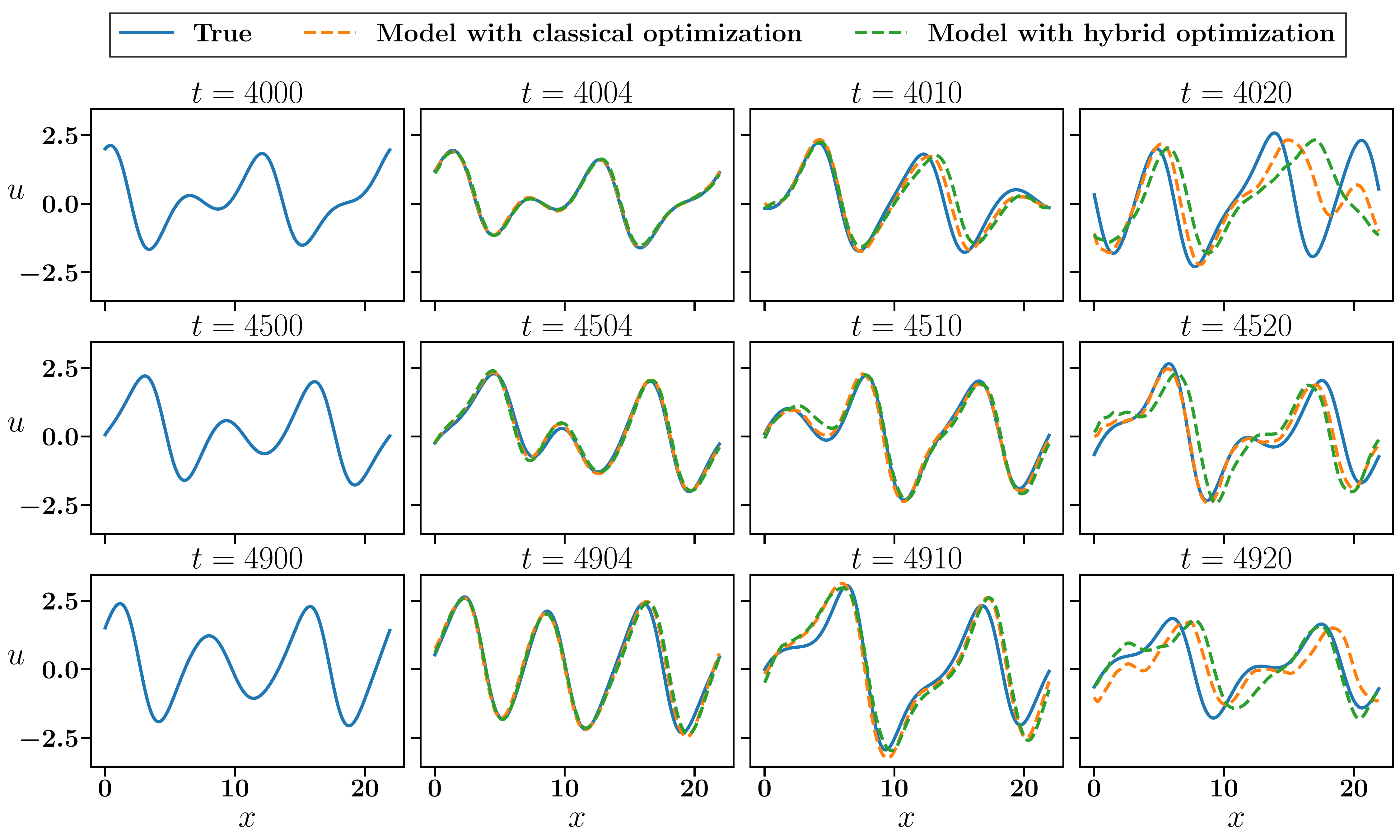}
    \caption{Solution profiles of the K-S equation for the true system, model trained with classical optimization, and model trained via hybrid optimization with different initial conditions in test data. The model with classical optimization are trained with 40 short-term batches with length of 2 time units from training data with resolution $dx=22/256, dt=2$. The model with hybrid optimization is further calibrated using the 20 sets of long-term statistics data. Both models are tested on the same resolution. The trained model via hybrid optimization corresponds to the point after one iteration in the $8^\mathrm{th}$ EKI in Fig.~\ref{fig:KSE_EKI_Error}.}
    \label{fig:KSE_profile_hybrid_2}
\end{figure}

\begin{figure}[H]
    \centering
    \includegraphics[width=0.5\textwidth]{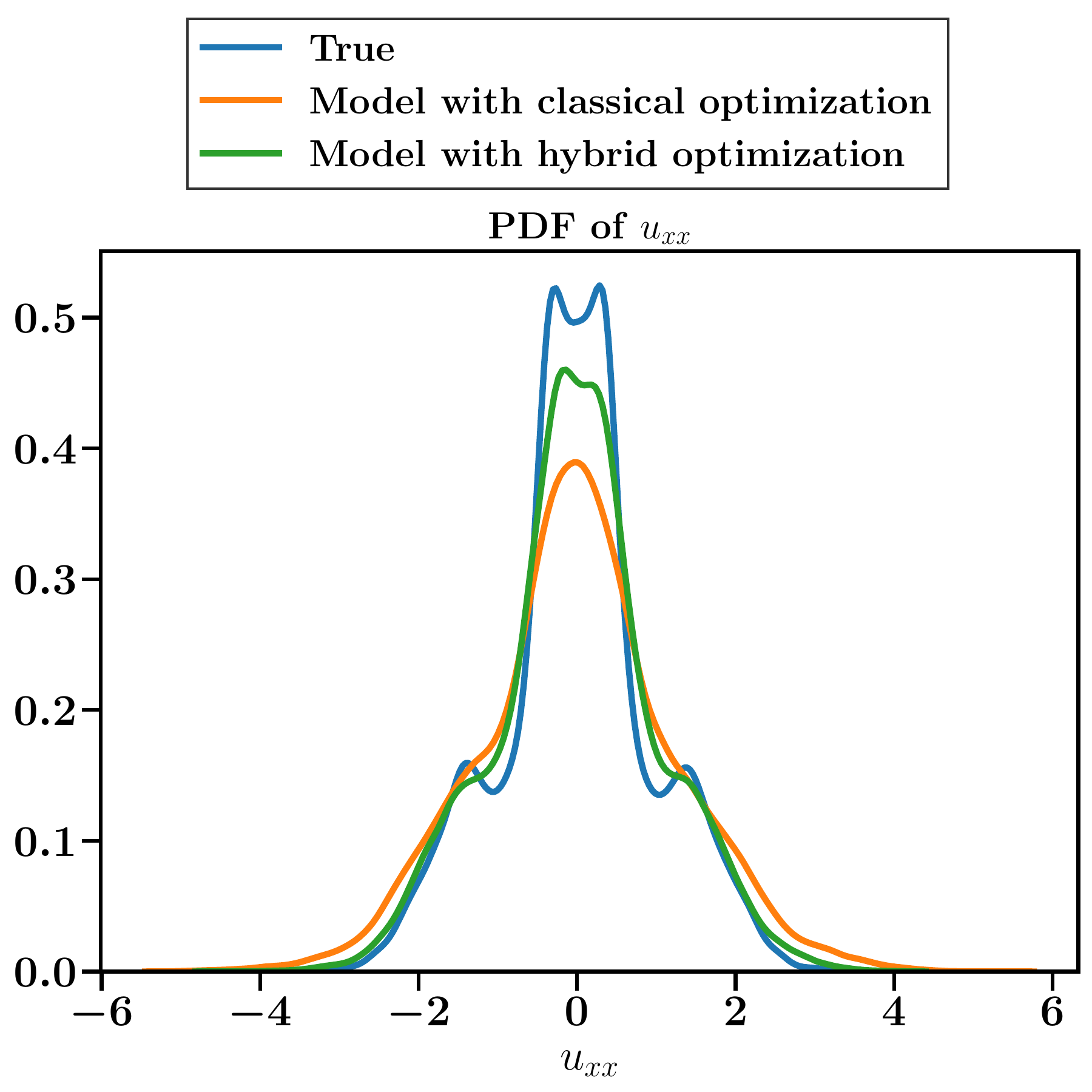}
    \caption{Probability density function of second spatial derivative $u_{xx}$ from long-term (1000 time units) simulation of true system, five sets of 200 time units simulation from model trained with classical optimization and model trained with hybrid optimization with initial conditions distributed evenly from test data. The trained model via hybrid optimization corresponds to the point after one iteration of the $8^\mathrm{th}$ EKI epoch in Fig.~\ref{fig:KSE_EKI_Error}.}
    \label{fig:KSE_uxx_hybrid_2}
\end{figure}

Figs~\ref{fig:KSE_profile_hybrid_1} to~\ref{fig:KSE_uxx_hybrid_2} present the results of trained models via the proposed hybrid optimization approach after one iteration of the $8^\mathrm{th}$ and $9^\mathrm{th}$ EKI epoch in Fig.~\ref{fig:KSE_EKI_Error}. Compared to the results of the trained model after two iterations of the $10^\mathrm{th}$ EKI epoch as presented in Figs.~\ref{fig:KSE_profile_hybrid} and~\ref{fig:KSE_uxx_hybrid}, we can confirm the robustness of the results obtained from the hybrid optimization approach.

\end{document}